\documentclass[letterpaper, 10 pt, conference]{ieeeconf}  

\IEEEoverridecommandlockouts                              

\usepackage{cite}
\usepackage{amsmath,amssymb,amsfonts}
\usepackage{algorithmic}
\usepackage{graphicx}
\usepackage{textcomp}
\usepackage{xcolor,colortbl}
\usepackage{subcaption}
\usepackage{booktabs}
\usepackage{hyperref}
\usepackage{comment}
\usepackage{bbm}
\usepackage{float}
\usepackage{dsfont}
\definecolor{Gray}{gray}{0.9}

\title{\LARGE \bf Lightweight Uncertainty Quantification with Simplex Semantic Segmentation for Terrain Traversability 
}
\author{Judith Dijk$^{1}$, Gertjan J. Burghouts$^{1}$, Kapil D. Katyal$^{2}$, Bryanna Y. Yeh$^{3}$, Craig T. Knuth$^{3}$, Ella Fokkinga$^{1}$, \\ Tejaswi Kasarla$^{4}$, Pascal Mettes$^{4}$
\thanks{*This work was partly supported by TNO Appl.AI program}
\thanks{$^{1}$ Judith Dijk, Gertjan Burghouts and Ella Fokkinga are with the department of Intelligent Imaging, TNO, 2597 AK, the Netherlands
        {\tt\small judith.dijk@tno.nl}}%
\thanks{$^{2}$Kapil Katyal is with Johns Hopkins University,
        Baltimore, MD, USA}%
\thanks{$^{3}$Bryanna Yeh and Craig Knuth are with the Johns Hopkins University Applied Physics Laboratory,
        Laurel, MD, USA
        {\tt\small bryanna.yeh@jhuapl.edu}}%
\thanks{$^{4}$Tejaswi Kasarla and  Pascal Mettes are with the University of Amsterdam, the Netherlands.} %
}

\begin{document}
\maketitle
\thispagestyle{empty}

\begin{abstract}
For navigation of robots, image segmentation is an important component to determining a terrain’s traversability. For safe and efficient navigation, it is key to assess the uncertainty of the predicted segments. Current uncertainty estimation methods are limited to a specific choice of model architecture, are costly in terms of training time, require large memory for inference (ensembles), or involve complex model architectures (energy-based, hyperbolic, masking). In this paper, we propose a simple, light-weight module that can be connected to any pretrained image segmentation model, regardless of its architecture, with marginal additional computation cost because it reuses the model’s backbone. Our module is based on maximum separation of the segmentation classes by respective prototype vectors. This optimizes the probability that out-of-distribution segments are projected in between the prototype vectors. The uncertainty value in the classification label is obtained from the distance to the nearest prototype. We demonstrate the effectiveness of our module for terrain segmentation. 
\end{abstract}
\section{Introduction}
Safe and efficient robot off-road navigation highly depends on accurate and actionable information of the environment. 
Semantic segmentation is a common component to determining the terrain's traversability with images (e.g. ~\cite{maturana}). This segmentation provides information about the current environment such as types of surfaces (e.g. puddles vs. dirt) or obstacles (e.g. tall grass vs. tree) that could impact robot navigation. If this segmentation model can run (almost) real-time on the robot, it can be used for path planning or replanning.


Not only the labels themselves are important, but also an estimate of their uncertainty. Such an uncertainty estimation enables online reasoning in path planning, e.g. uncertain areas can be avoided or entered more carefully. Such an approach is e.g. shown by Cai et al.~\cite{cai2022probabilistic} where traversability estimates obtained by traction parameters of the platform are used for risk estimation, and by Hakobyan et al.~\cite{hakobyan2019risk} who proposed risk-aware motion planning and control Using conditional value-at-risk (CVaR)-Constrained Optimization. 


Standard methods for semantic segmentation focus on label accuracy and not on the accuracy of the uncertainty estimate. Segmentation models that provide uncertainty are not optimized for robotics and embedded scenarios~\cite{atigh2022hyperbolic,jungo2019assessing}, where uncertainty quantification needs to be fast, compact, and without impacting the segmentation accuracy.

The approach presented in this paper consist of a simple, lightweight module for uncertainty estimation for image segmentation. This module, dubbed Simplex Semantic Segmentation, can be connected to any pretrained semantic segmentation model as it is architecture agnostic. Our approach is based on the prototypes approach that was developed for classification, often on imbalanced datasets with rare classes~\cite{rebuffi2017icarl, cen2021deep, kasarla, liu2018learning, mettes2019hyperspherical, zhou2022all}. 
In this paper, we apply this prototype approach to obtain the uncertainty for semantic segmentation, where each {\em pixel} of an input image needs to be labeled. 


This paper is organized as follows: Section~\ref{chapter:segmentation_uncertainty_estimation} presents our method for the prototype module and the uncertainty estimation.  
In Section~\ref{chapter:experiments}, these proposed methods are evaluated. In Section~\ref{chapter:conclusions}, the findings are summarized and discussed.

\section{Proposed approach}
\label{chapter:segmentation_uncertainty_estimation}
\begin{figure*}
     \centering
     \includegraphics[width=0.98\textwidth]{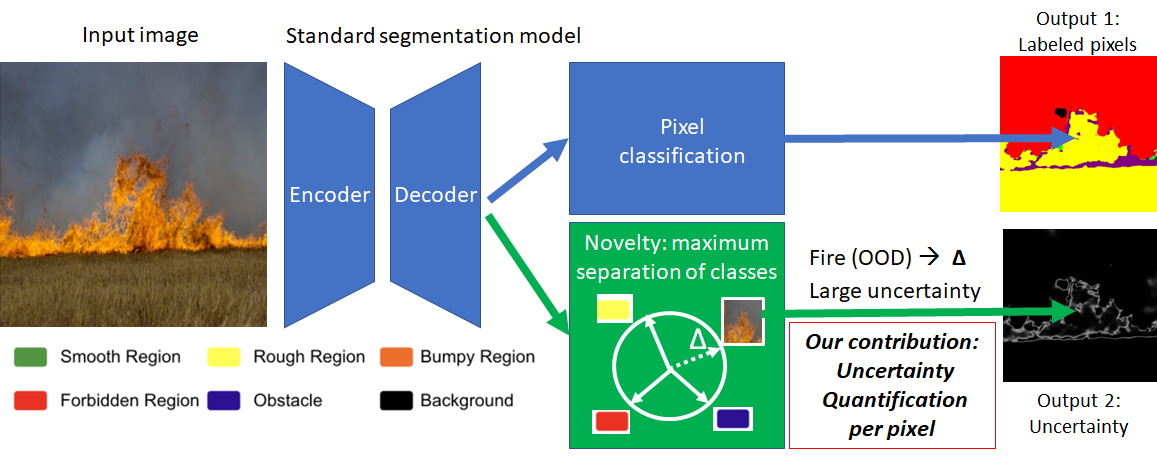}
     \caption{This figure describes our ability to predict segmentation classes with uncertainty. The input image consists of regions that are in-distribution and areas that are out-of-distribution. We are able to extend a traditional segmentation model (a) by estimating uncertainty through distances to the nearest prototype vector (b). In this example, we are able to detect out-of-distribution areas (fire) through our uncertainty estimation.
    }
     \label{fig:method}
\end{figure*}
\subsection{Rationale}
To estimate the uncertainty of image segmentation, we propose a simple, light-weight module that can be connected to any pretrained segmentation model. The module only needs a feature map and is independent of the underlying architecture. As the  module is only an extra model head, it only adds a marginal  computation cost. 
Our module builds on the prototypes approach, that is commonly used to improve classification  of imbalanced datasets with rare classes~\cite{rebuffi2017icarl, cen2021deep, kasarla}. In the standard prototype approach, each segmentation class is represented by its respective prototype vector.
In our proposed method, we apply this prototype approach to obtain the uncertainty for semantic segmentation, where each {\em pixel} of a given input image is labeled to a specific class. 
 In the proposed method, we apply this prototype approach to obtain the {~\em uncertainty} the labeled pixel. During training, these prototypes are incorporated to maximize distance between classes. In the inference phase, samples are then classified based on their distance to prototypes. 
The prototype approach maximizes the probability that pixels of unknown image segments will be projected in the void space between the unknown classes. The uncertainty is inferred from the distance to the nearest prototype vector. Figure \ref{fig:method} provides an overview of our approach and how the proposed module is connected to any segmentation model.

\subsection{Method}
The starting point of our approach is a given pretrained image segmentation model $h$: $\mathcal{X}$ $\rightarrow$ $\mathcal{Y}$ which classifies image pixels $x_n$ into $N$ respective classes $y_n$ using a set of classification parameters $W_c$. We rewrite $h(\cdot; W_c) = g(f(\cdot; W_f); W_g)$, where $f(\cdot; W_f)$ is the model's backbone which outputs a feature map, and $g(\cdot; W_g)$ is the pixel classification head that transforms the feature map into class predictions. $W_f$ and $W_g$ are the parameters connected to the functions $f$ and $g$, respectively. 
Our objective is to predict the uncertainty of the labels using parameters $W_u$ by the module 
$U = u(f(\cdot; W_f); W_u)$, with $W_u$ the uncertainty estimation parameters. 
Note that this uncertainty function U can be applied to any spatial feature map $f$, without loss of generality, which makes our approach applicable to any segmentation model.

The uncertainty function $U$  can be further separated using  prototype vectors $P$
\begin{equation}\label{eq:u}
U= m(l(\cdot, W_u)\,\cdot\,P)
\end{equation}
where $m$ is a fixed mapping and $l(\cdot, W_u)$ a neural network that projects pixel features onto the respective class prototype to maximize the output value for the pixel's correct segmentation class \cite{kasarla}. 

The prototype vectors are maximally separated on the $(N-1)$ dimensional hypersphere. The derivation is recursive \cite{mathexchange}:
\begin{align}
P_1&=\begin{pmatrix}1&-1\end{pmatrix} \; \; \in \; \mathbb R^{1\times2}\\
P_k&=\begin{pmatrix}1&-\frac{1}{k}\mathbf1^T\\ \mathbf0&\sqrt{1-\frac1{k^2}}\,P_{k-1}\end{pmatrix} \; \in \; \mathbb R^{k\times(k+1)}
\end{align} 
with $\mathbf0$ and $\mathbf1$ respectively the $0$ and $1$ column vectors. The columns of $P_k$ are $k + 1$ equidistant vectors on the unit sphere in $\mathbb{R}^k$. With $N$ classes, the prototypes are obtained by constructing $P_{N-1}$, yielding $N$ vectors in $N - 1$ dimensions. The rationale is that if the pixel's projection is further away from the prototype vector, it is more uncertain. 

The uncertainty is inversely proportional to the maximum of a pixel's scores across the class prototypes: $m(\textbf{x}) = 1 - \sigma(max(\textbf{x}))$, where $\textbf{x}$ are the pixel's prototype scores from $l(\cdot, W_u)\,\cdot\,P$, and $\sigma$ is the softmax operator. Note that this uncertainty is not calibrated yet.

To predict uncertainty, the module needs to learn the classes first. Therefore, the learning objective is to align the outputs of $l(f(x_i), W_u)$ with the class prototypes $P_{y_i}$, with $x_i$ and $y_i$ from the training set $\mathcal{S}$, using the cosine loss: 
\begin{equation}\label{eq:mapping}
\mathcal{L} = - \sum_{i=1}^{M} \frac{m(l(f(\mathbf{x}_i), W_u)) \cdot P_{y_i}}{\lVert m(l(f(\mathbf{x}_i), W_u)) \rVert \cdot \lVert P_{y_i} \rVert}
\end{equation}

After learning to project a pixel's feature onto the prototype vector, at inference we can tell if a test pixel is projected far away from such a vector. The further away, the more uncertain the pixel's classification. We hypothesize that this works best if the prototype vectors are maximally apart.


\section{Experiments}
\label{chapter:experiments} 

\subsection{Base dataset}
Since we target off-road applications, we use a base dataset recorded in off-road environments with a  diverse range of objects and terrain types. From the online available  datasets~\cite{wigness2019rugd, rellis, mayuku2021multi} we choose 
Rellis3D~\cite{rellis}, which is collected in an off-road environment with 6,235 annotated images. The classes in this dataset are Concrete, Asphalt, Gravel, Grass, Dirt, Sand, Rock, Rock Bed, Water, Bushes, Tall Vegetation, Trees, Poles, Logs, Void, Sky, Sign. 
The environment is complex and differs between images.  To segment the images in terms of traversability, we adopt the six classes from \cite{ganav}: 
Smooth, Rough, Bumpy, Forbidden, Obstacles and Background. 

\subsection{Model training}
For the experiments, we select DeepLabV3+ \cite{deeplabv3plus} for its simplicity and broad usage. DeepLabv3+ contains a decoder that refines the segmentation using atrous convolutions and a spatial pyramid pooling. As a backbone, we select a Resnet50 \cite{resnet} pretrained on ImageNet \cite{imagenet}, as this is one of the most commonly used backbones. The images are resized to 512 $\times$ 512 pixels and augmented with standard transformations: horizontal flip, shift, scale, rotate and color jitter (all at a probability of 0.5). The batch size is 4. The model is trained for 25 epochs, at a learning rate of 0.001. All weights, including the backbone, are optimized during training. The segmentation results and accuracy of the segmentation are presented in the Appendix. 

\subsection{Uncertainty estimation}

For the uncertainty estimation, we start by comparing the uncertainty values measured in the test set of Rellis3D and the uncertainty values measures in images from other datasets. The uncertainty values should be higher for datasets that are different. For simplicity, we assume that all pixels in an image are certain for Rellis3D and uncertain for other datasets, because for these datasets we do not have detailed annotations of certain or uncertain image segments. This assumption is often violated, because images from the other datasets may have segments that are very similar to Rellis3D.
We compare against the DeepLabV3+ baseline, which is the same model, but without our uncertainty module. Its outputs are transformed into uncertainty values by the mapping $m(\cdot)$ from Equation \ref{eq:mapping}. This methods is referred to a {\em standard method} in the remainder of this paper. 

The uncertainty performance is measured by the 1) Receiver Operator Characteristic (ROC) curve (graph), in which the true positive rate is plotted against the False Positive Rate  and the 2) Area Under the Curve (AUC). The best performance will be found if the left upper corner is reached, in which case the AUC is 1.

The AUC scores for both methods for all datasets is shown in Table~\ref{table:uncertainty_results}. The ROC curves can be seen in Figure~\ref{fig:roc} in the Appendix. On all datasets, our method shows a better uncertainty estimation performance than the standard method. On CUB-200, MS-COCO and KITTI, the performance is very similar with the standard method. These datasets are reasonably different from Rellis3D, but often contain vegetation, streets, humans and cars, which are also in Rellis3D. For WiderPerson and Fukuoka, our method is favorable, which can be explained by respectively the different viewpoint (aerial) and environment (indoor). The most prominent insight is that our method is uncertain about fog and fire, whereas the standard method is as certain as it is for Rellis3D on which it was trained. This result demonstrates the merit of our method for uncertainty estimation in practical cases.

\begin{table}
\centering
\caption{AUC for the standard method and the proposed method. The datasets are ordered from low to high difference with Rellis3D.}
\label{table:uncertainty_results}
\begin{tabular}{ l l c c }
\toprule  
    Dataset & Contents & standard  & our  \\
    & & method & method \\
    \midrule
    
    MS-COCO\cite{mscoco}& common objects & 0.766 & \cellcolor{Gray} 0.794 \\ 
    
    WiderPerson~\cite{widerperson}& aerial  & 0.796 & \cellcolor{Gray} 0.846 \\ 
    KITTI~\cite{behley2019semantickitti} & self-driving cars & 0.853 & \cellcolor{Gray} 0.859 \\
    SceneParse150~\cite{sceneparse150} & common segments & 0.770 & \cellcolor{Gray} 0.861 \\
    
    Fukuoka~\cite{fukuoka}& indoor robot &  0.847 & \cellcolor{Gray} 0.904 \\
    
    CUB-200\cite{cub200}&  birds & 0.841 & \cellcolor{Gray} 0.856 \\
    
    Fog and Fire & outdoor anomalies & 0.478 & \cellcolor{Gray} 0.881 \\ 
   \bottomrule
\end{tabular}
\vspace{-4mm}
\end{table}

\subsection{Segment-specific uncertainty}

\begin{figure*}
     \centering
     \includegraphics[width=0.78\textwidth]{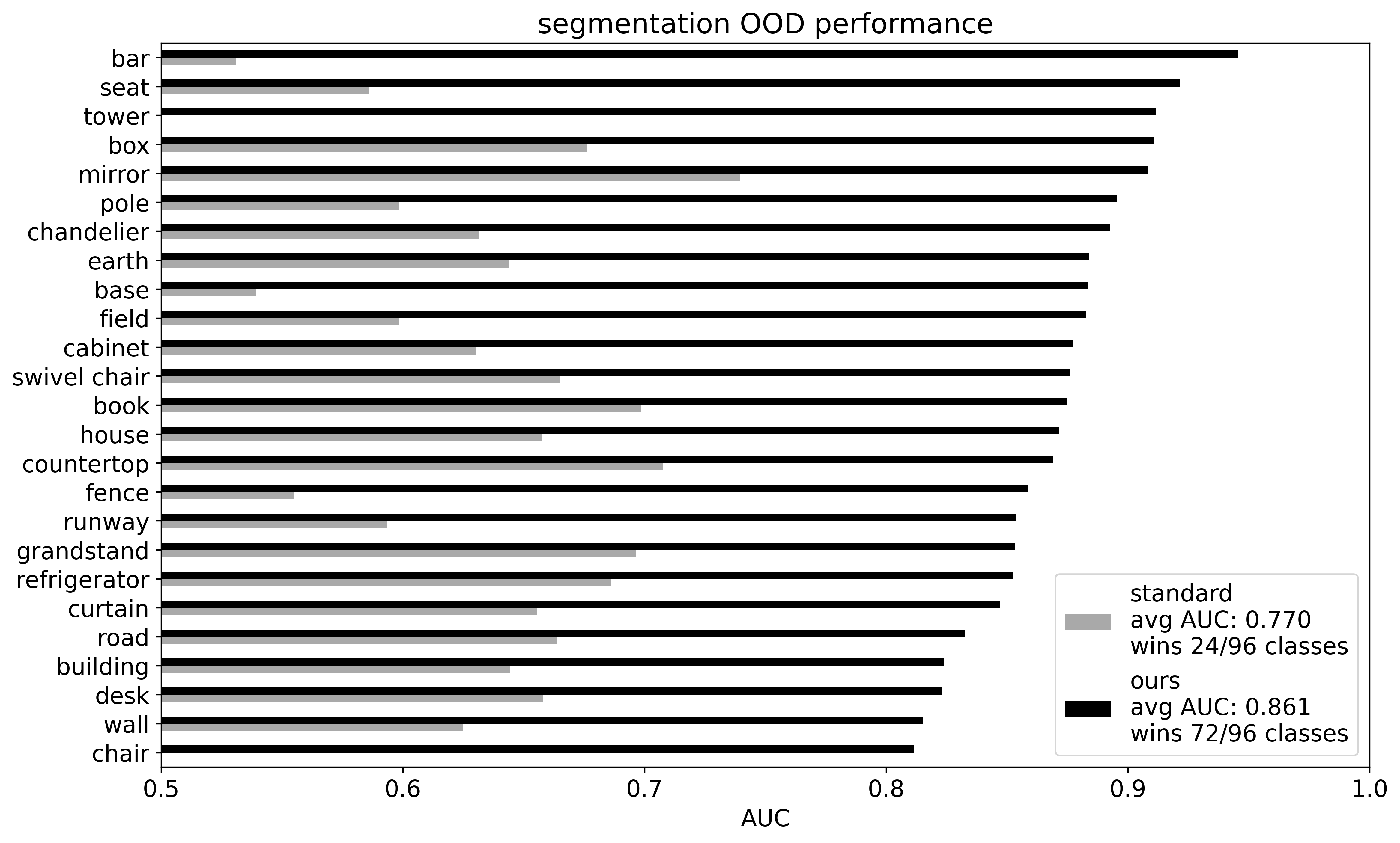}
     \caption{Segments from SceneParse150 for which our model (trained on Rellis3D) assigns the highest uncertainties.}
     \label{fig:ood_ranks}
\vspace{-3mm}
\end{figure*}

We evaluate the uncertainty estimation for respective segment classes across images. In the SceneParse150~\cite{sceneparse150} dataset, most segmentation classes are semantically and visually different from the Rellis3D classes. The performance is measured by the AUC on uncertainty values per segment class in comparison to Rellis3D uncertainty values. The segmentation classes that are most uncertain are shown in Figure~\ref{fig:ood_ranks}. Indeed, these classes are most different from Rellis3D, hence are expected to yield high uncertainties. In comparison to the standard method, our method yields higher uncertainty values for the segmentation classes that deviate from the training dataset.

\begin{figure}
\centering
\begin{tabular}{c c c}
    & uncertainty & uncertainty \\
    test image & ours & standard \\        
     \includegraphics[width=0.13\textwidth]{figures/OOD_pigs_in.png}
 &
        \includegraphics[width=0.13\textwidth]{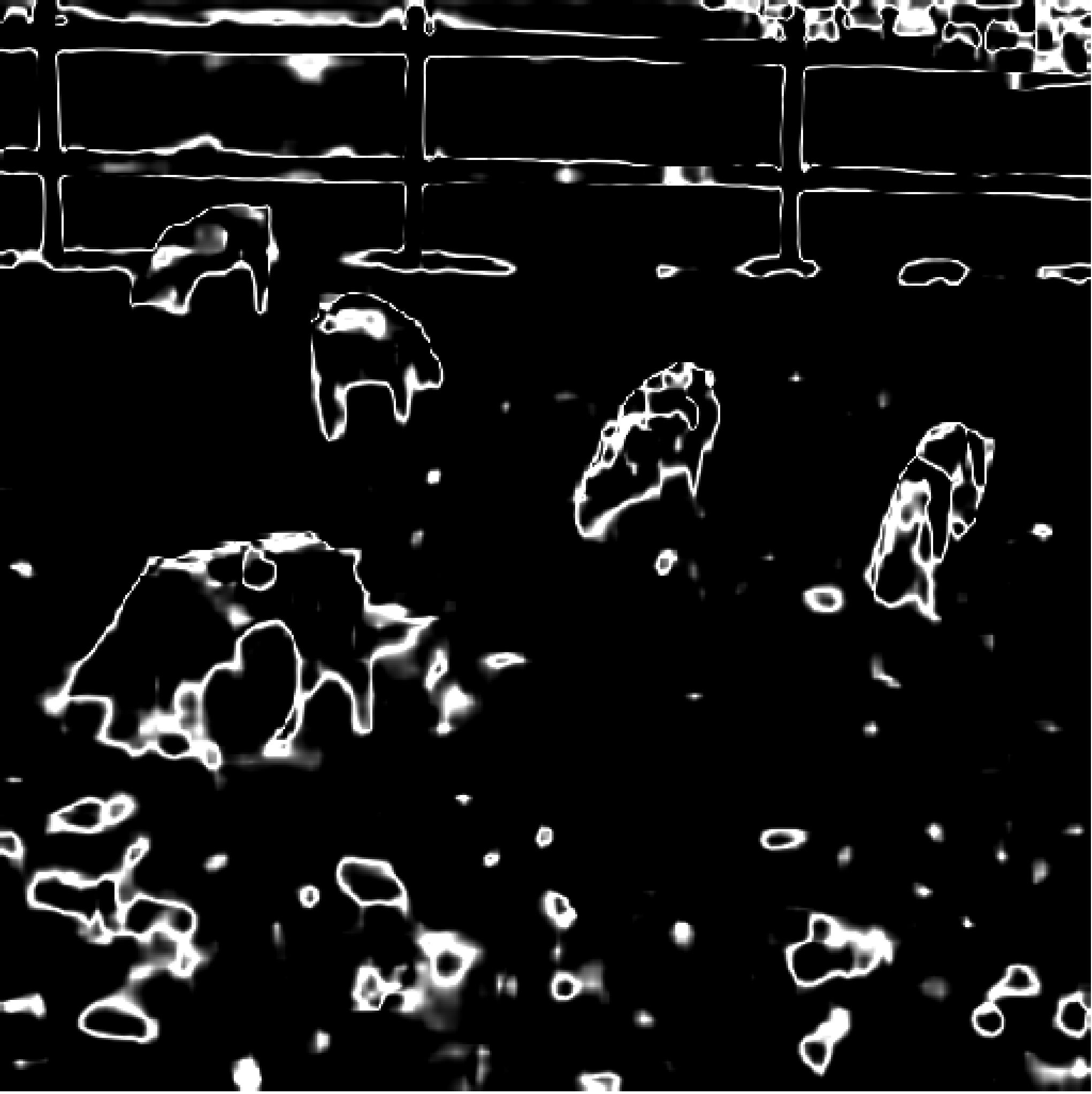}
    &
        \includegraphics[width=0.13\textwidth]{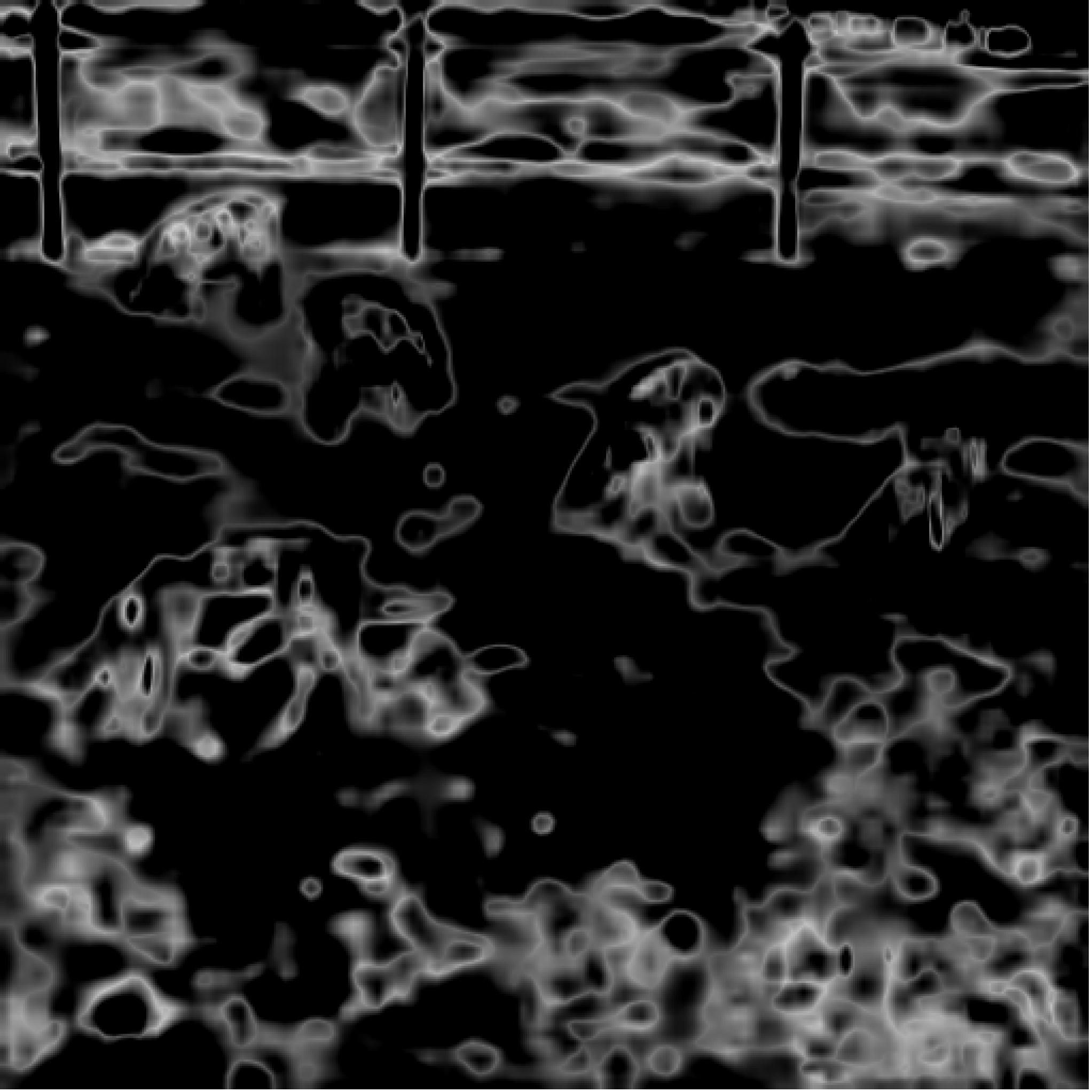}\\
    \includegraphics[width=0.13\textwidth]{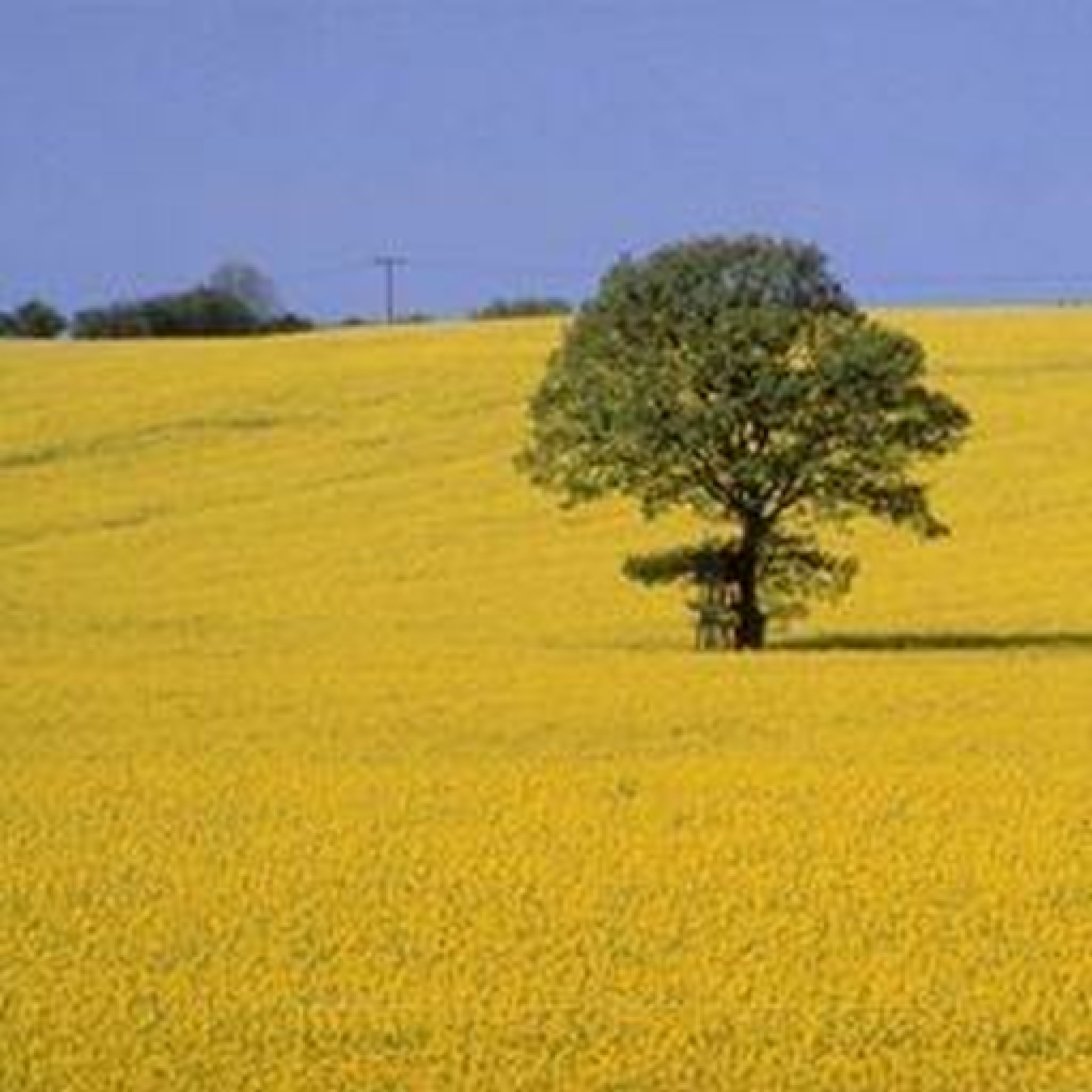}
    &
        \includegraphics[width=0.13\textwidth]{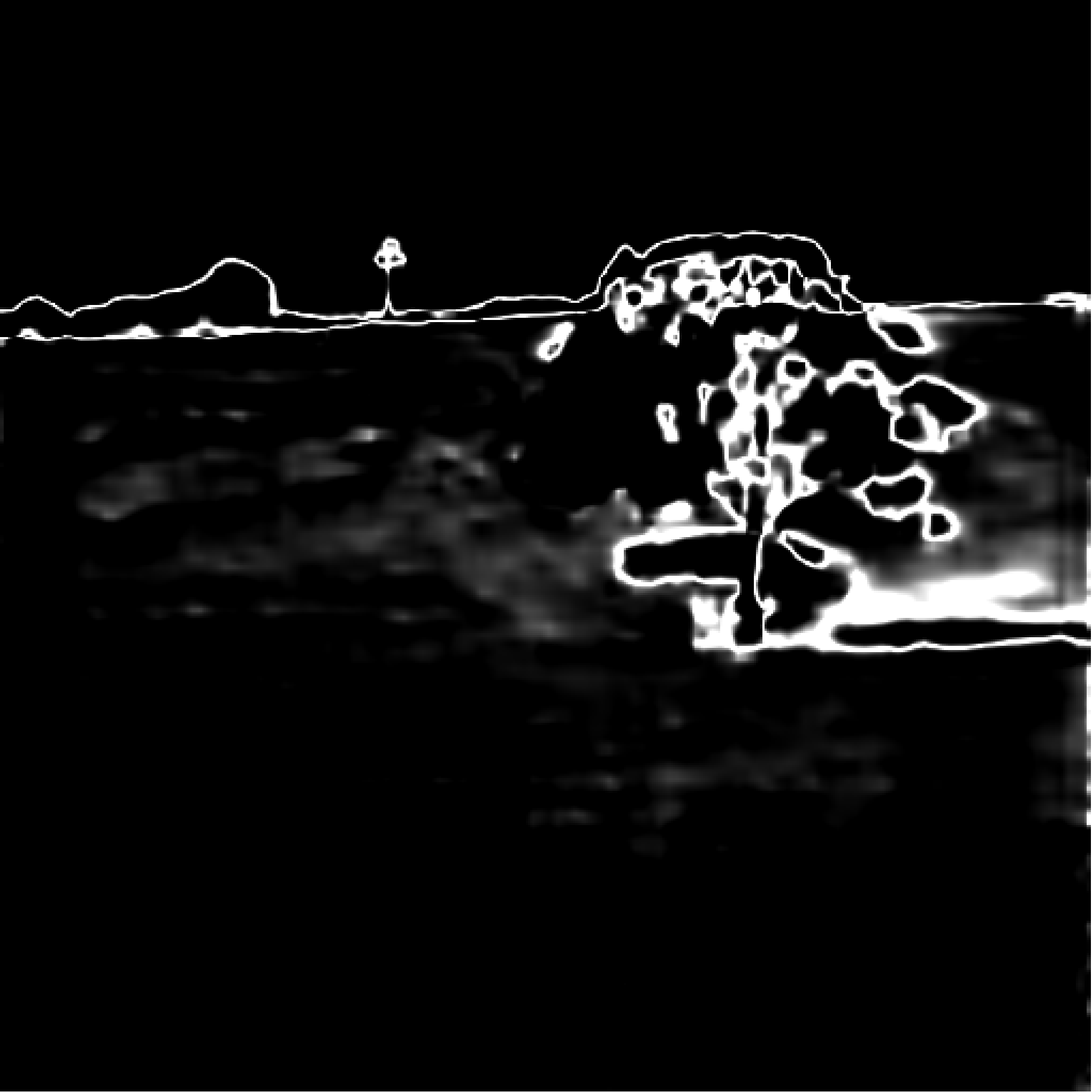}
    &
        \includegraphics[width=0.13\textwidth]{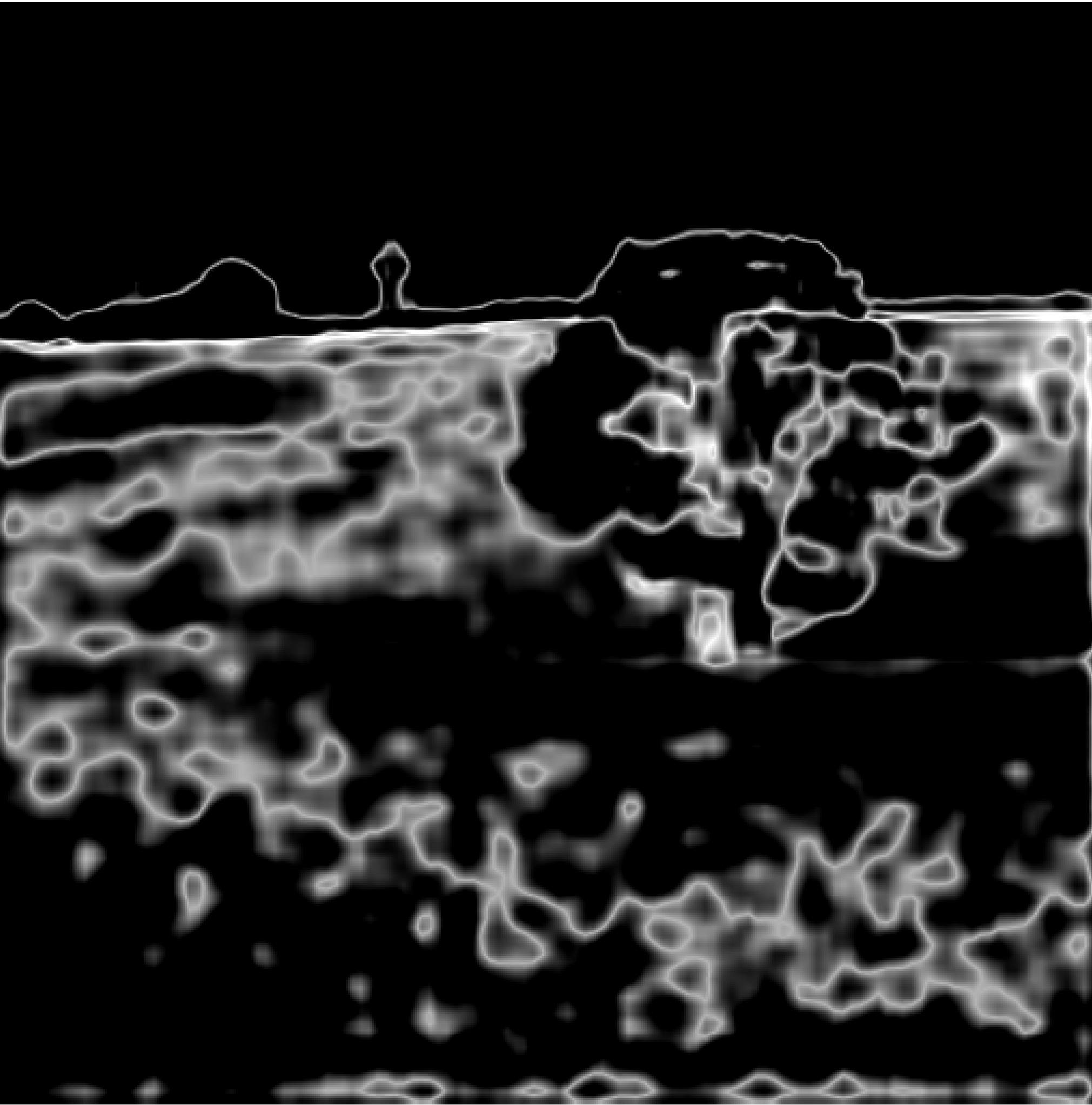}
    \\

   \end{tabular}
       \caption{Our model  (middle) yields high uncertainty values for the tree, pigs and fence, whereas the uncertainty  obtained with the standard method is scattered (right).\label{fig:ood_pigs_paper}}
\vspace{-4mm}
\end{figure}

\subsection{Uncertainty visualization}
It is also possible to show the uncertainty per pixel. An example is shown in Figure \ref{fig:method}. Here the {\em Input image} shows both in-domain (vegetation, sky) and out-of-domain (fire) regions. {\em Output1} shows the segmentation and {\em Output2} shows the resulting uncertainty values: low (black) for the in-domain image segments and high (white) for the out-of-domain image segments. This is the desired behaviour: the robot can make navigation decisions about the image segments where it is certain, while operating in safe mode at the uncertain segments. It can be seen that the more uncertain values are found around the edges of segments, which is as expected as these pixels might contain information from both of these segments, and the labeling  will often be based on surrounding pixels. This can also be seen in Figure~\ref{fig:ood_pigs_paper} for two images from the SceneParxe10 dataset, where the uncertainty found for the unknown class of the pigs is much higher with our proposed method than with the standard method.  
More examples are presented in Figure~\ref{fig:out_domain} in the Appendix. In the bottom image, it can also be seen that the uncertainty for the field is much smaller for our method than for the standard method. This smaller uncertainty will allow for better path planning and faster navigation, by avoiding paths with high uncertainty. 

\subsection{Inference computational costs}
Regarding the computational cost of inference on a test image, our method is advantageous. The most common method for uncertainty evaluation is to use Monte-Carlo dropout \cite{dropout}. Although very effective, this requires a number of repeated feed-forward calculations of the model with randomly sampled weight parameters, which needs much computation power and can cause long latency. In our experiments, the model is DeepLabV3+ which has 11.9M weights. Instead of running this model multiple times, our module is low-cost and has to be run only once. For six traversability classes, our module has 1285 weights, i.e. 0.01\% of the total model. These weights are required to transform the model's feature map (the final upscaling layer), a tensor of 512 $\times$ 512 $\times$ 256 dimensions, into 512 $\times$ 512 $\times$ 5 dimensions. This transformation is $l(\cdot, W_u)$ in Equation \ref{eq:u}. It is implemented as a channel-wise convolution from 256 to 5 dimensions. The mapping P from Equation \ref{eq:u} projects the output of $l(\cdot, W_u)$ from 5 to 6 dimensions, which are the 6 traversability classes. This projection is a fixed matrix multiplication, no learnable weights, implemented as a highly efficient vectorized operation.
\section{Conclusions and discussion}
\label{chapter:conclusions}
We propose an extra, lightweight module for a semantic segmentation network, which provides a high quality segmentation label per pixel and an uncertainty estimate for these labels. The module first optimizes the segmentation by maximizing the separation of the different classes in the training phase. In the inference phase, the uncertainty of a pixel can be estimated as the distance of the current pixel to the center of the class it is labeled to. We have shown that our approach performs
on par (with GA-Nav-r8) or just a little (for DeepLabV3++) better than standard models. 
We have shown high uncertainty values for data that really differs from the Rellis3D data we trained on. This indicates that  the added module provides a good uncertainty estimate for pixels and segments in an image.

In future work, 
We will evaluate the uncertainty estimation in a more quantitive way by estimating the 'ground truth uncertainty' using monte carlo estimations. Next to that, we will compare our approach against existing uncertainty estimation approaches. We will also focus on on calibrating the uncertainty. 
The uncertainty visualisation shows that our labeling is more uncertain on edges in the image. When using the labeling and uncertainty estimation for navigation, this means that these areas can be avoided when possible.
In future work we will implement this simplex semantic segmentation and uncertainty approach for robust long-distance navigation on a physical robot, both for providing trusted generalized features for self-supervised traversability prediction as in ~\cite{polevoy2021complex} and informing risk-based planning such as CVaR-Conditional optimization mentioned earlier. This will provide information on how well this uncertainty can actually be used for navigation purposes.

\newcommand{\kapil}[1]{\textcolor{cyan}{#1}}

\bibliographystyle{IEEEtran}
\bibliography{sections_review/refs}
\section*{Appendix: Illustrations}

In this appendix, we show some figures illustrating the results presented earlier.

\begin{figure*}
    \centering
    \begin{tabular}{c c c c c}
    & & & uncertainty & uncertainty \\
    test image & ground truth & predicted & ours & standard \\
         \includegraphics[width=0.175\textwidth]{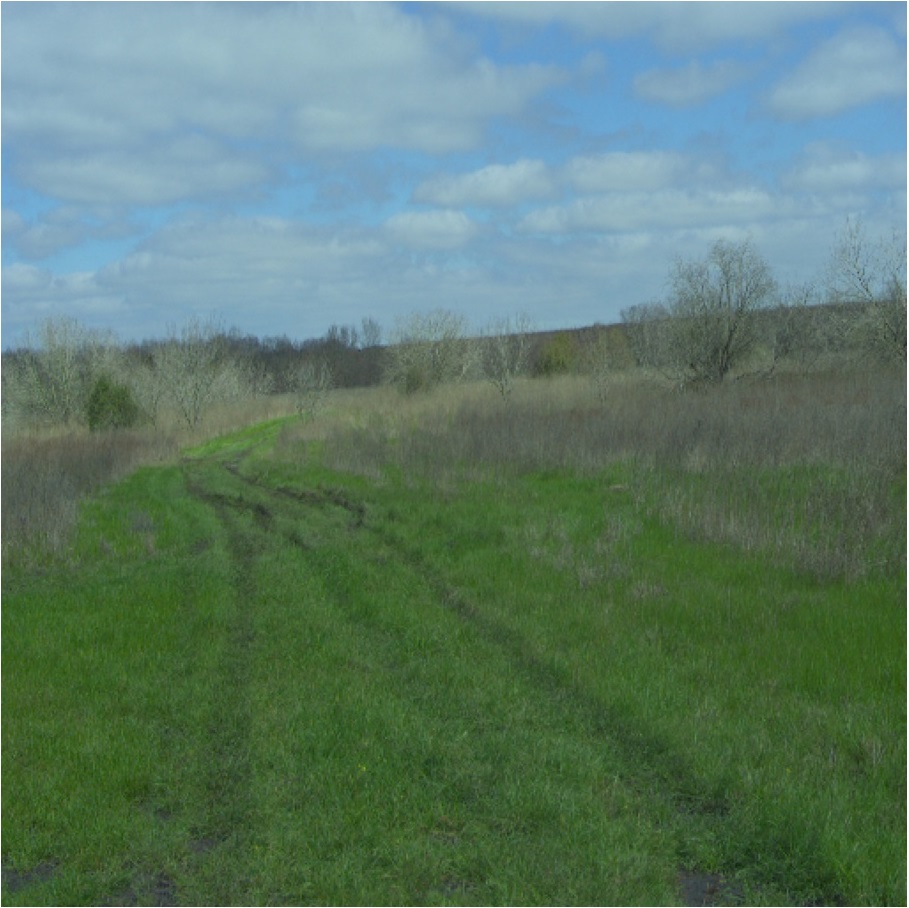}    &
        \includegraphics[width=0.175\textwidth]{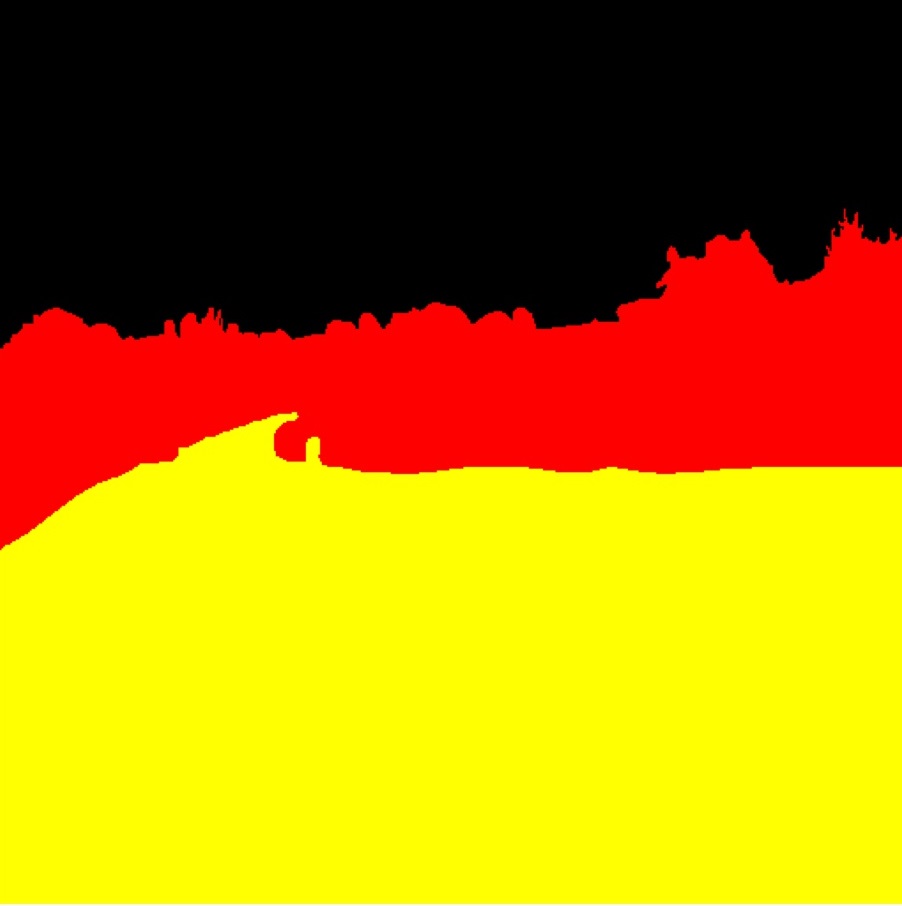}      &      \includegraphics[width=0.175\textwidth]{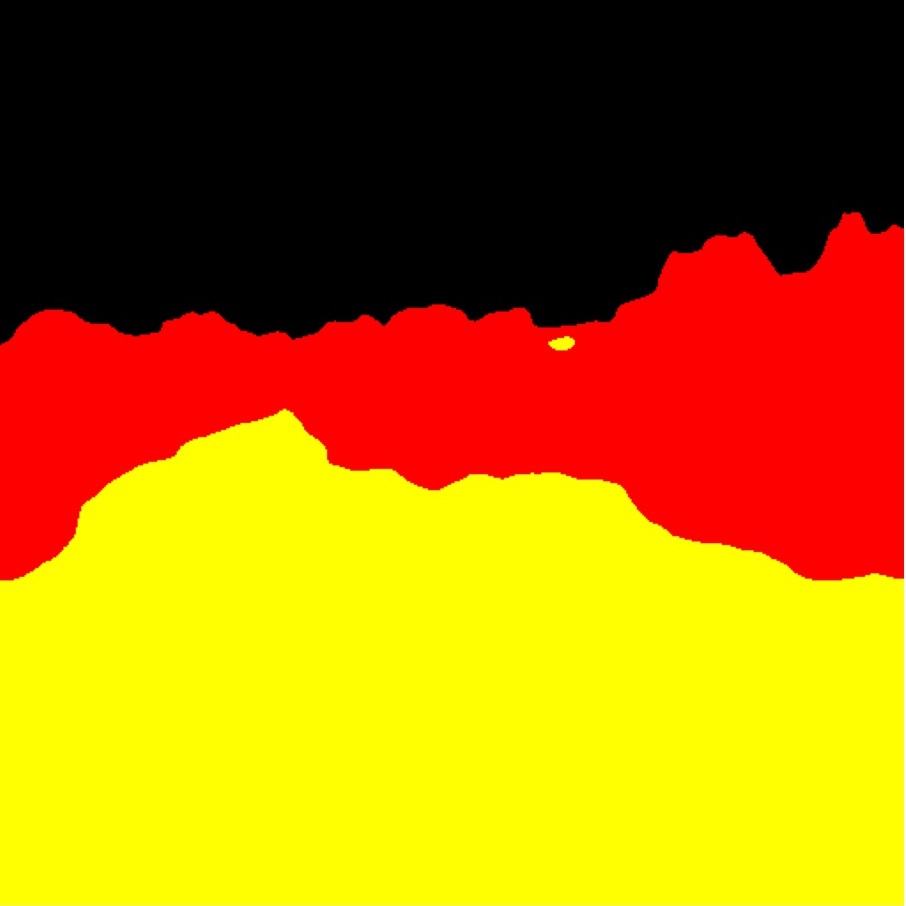}&
        \includegraphics[width=0.175\textwidth]{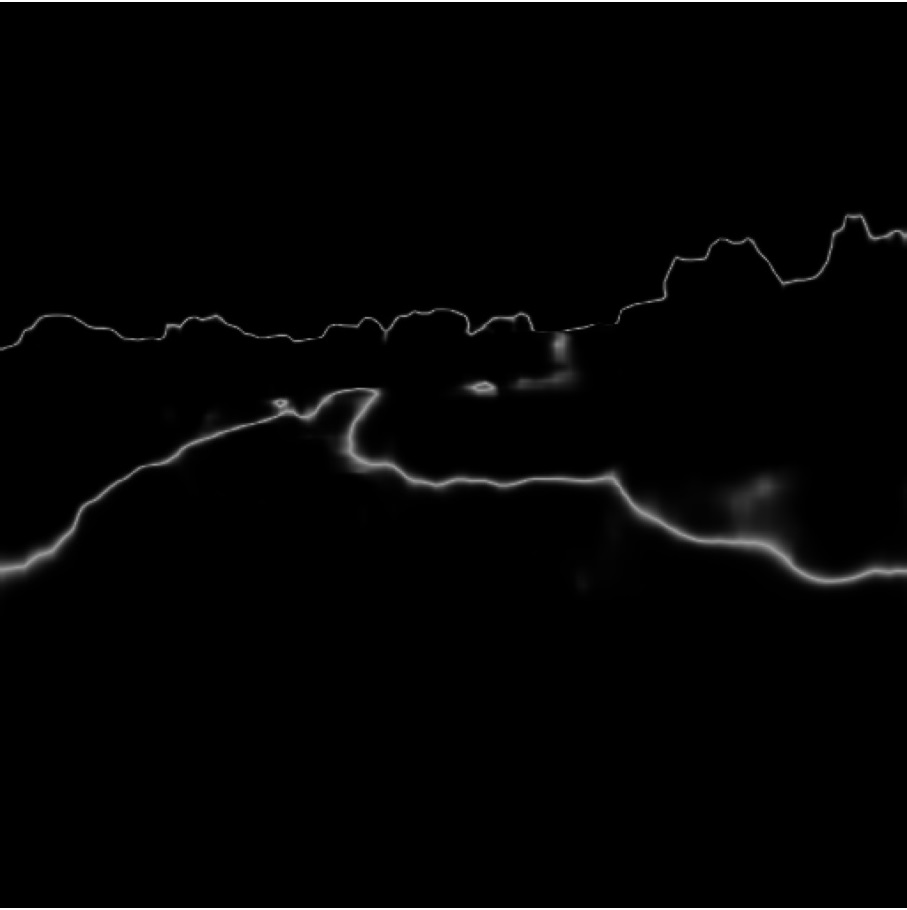}&
         \includegraphics[width=0.175\textwidth]{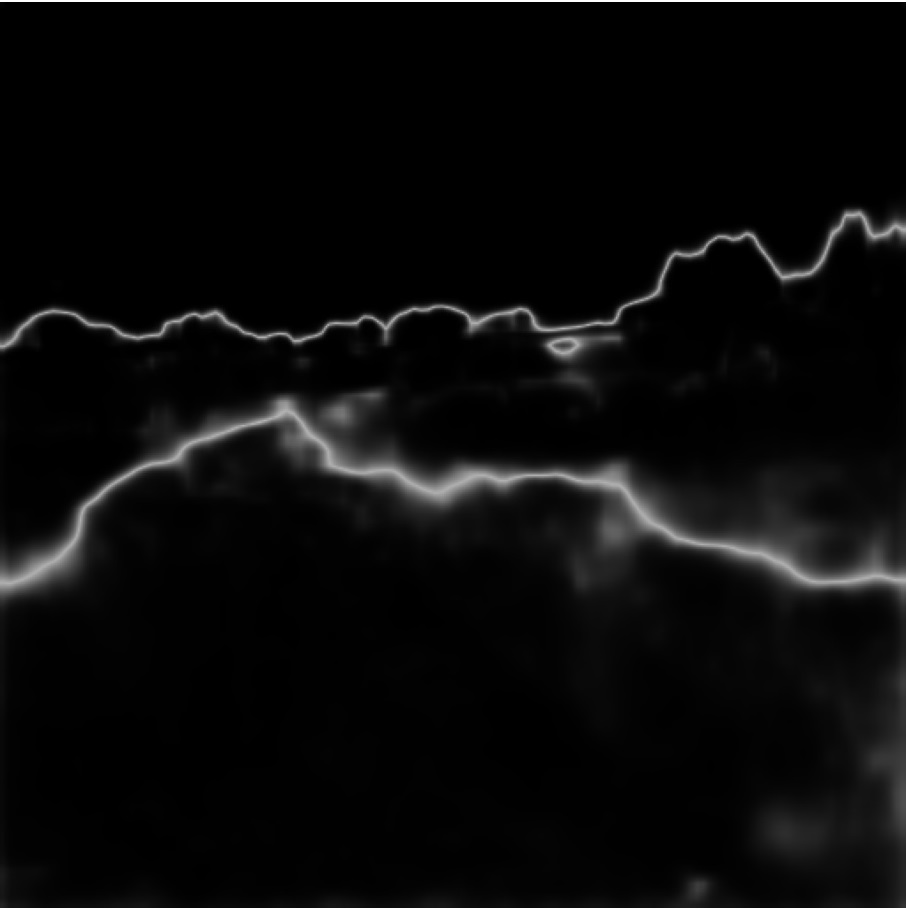} \\
        \includegraphics[width=0.175\textwidth]{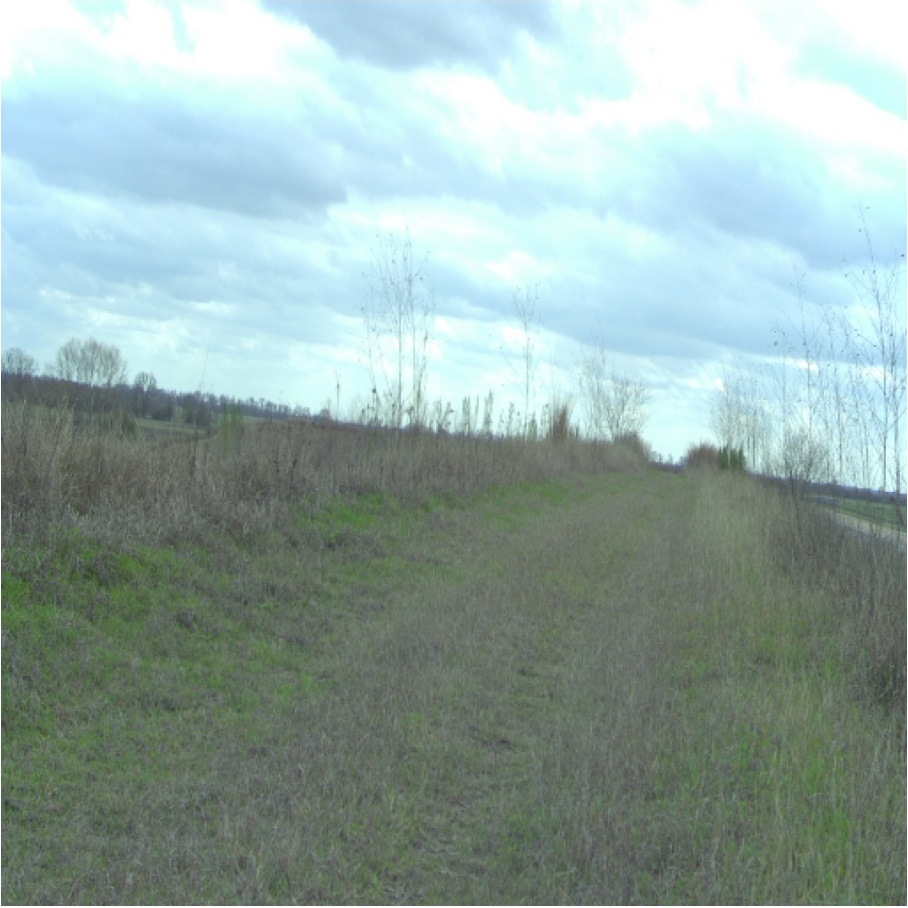} &
         \includegraphics[width=0.175\textwidth]{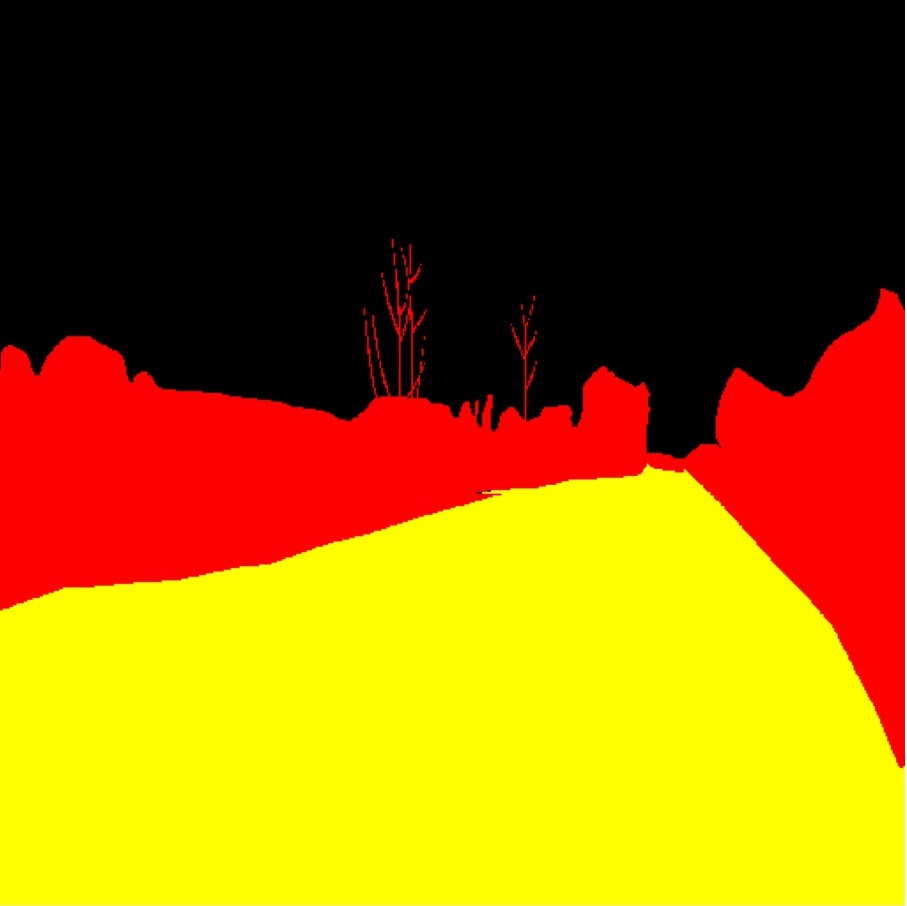} &
         \includegraphics[width=0.175\textwidth]{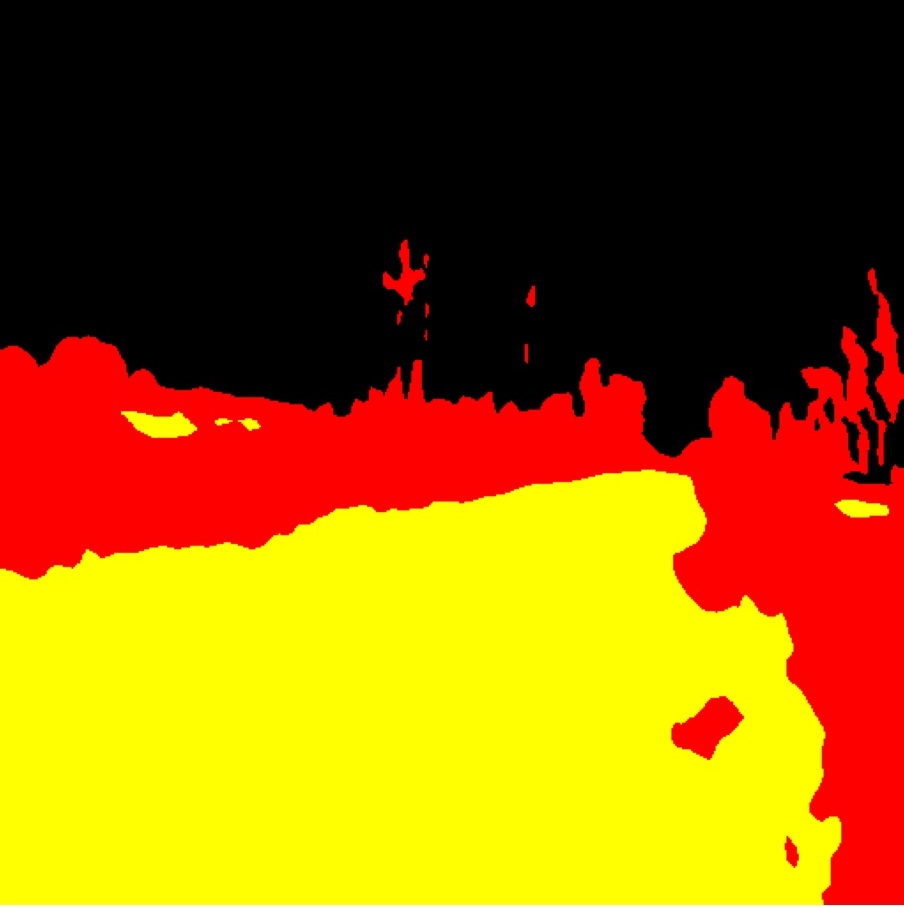} &
        \includegraphics[width=0.175\textwidth]{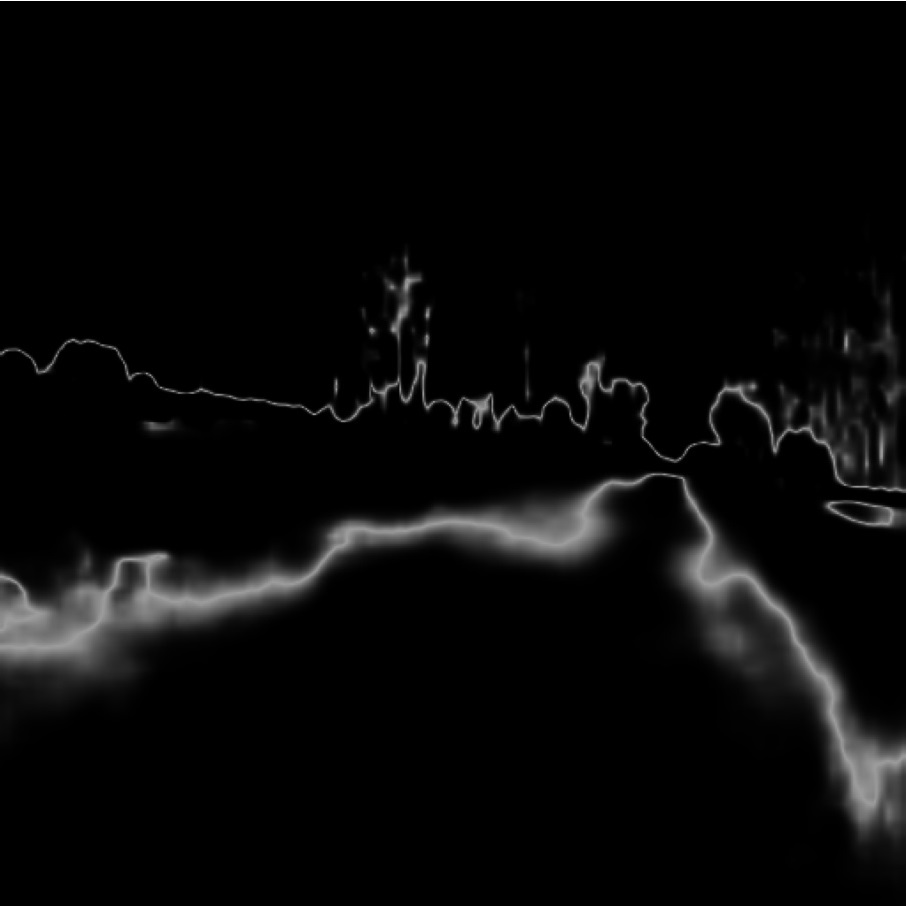} &
        \includegraphics[width=0.175\textwidth]{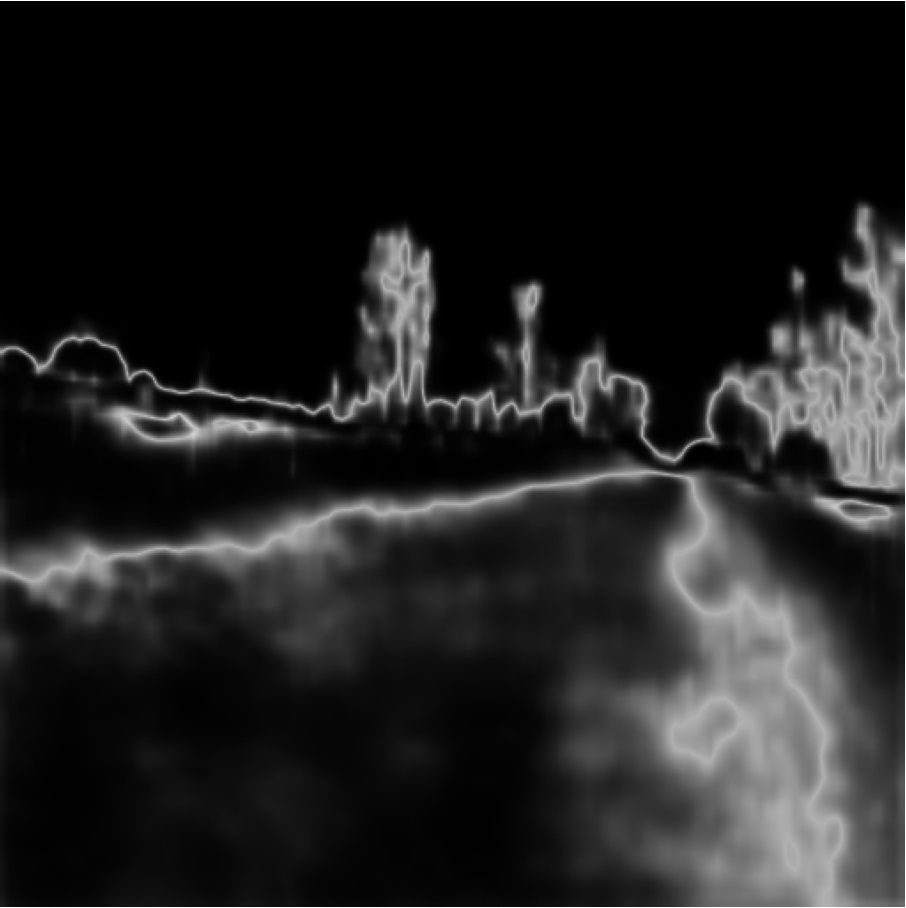} \\
        \includegraphics[width=0.175\textwidth]{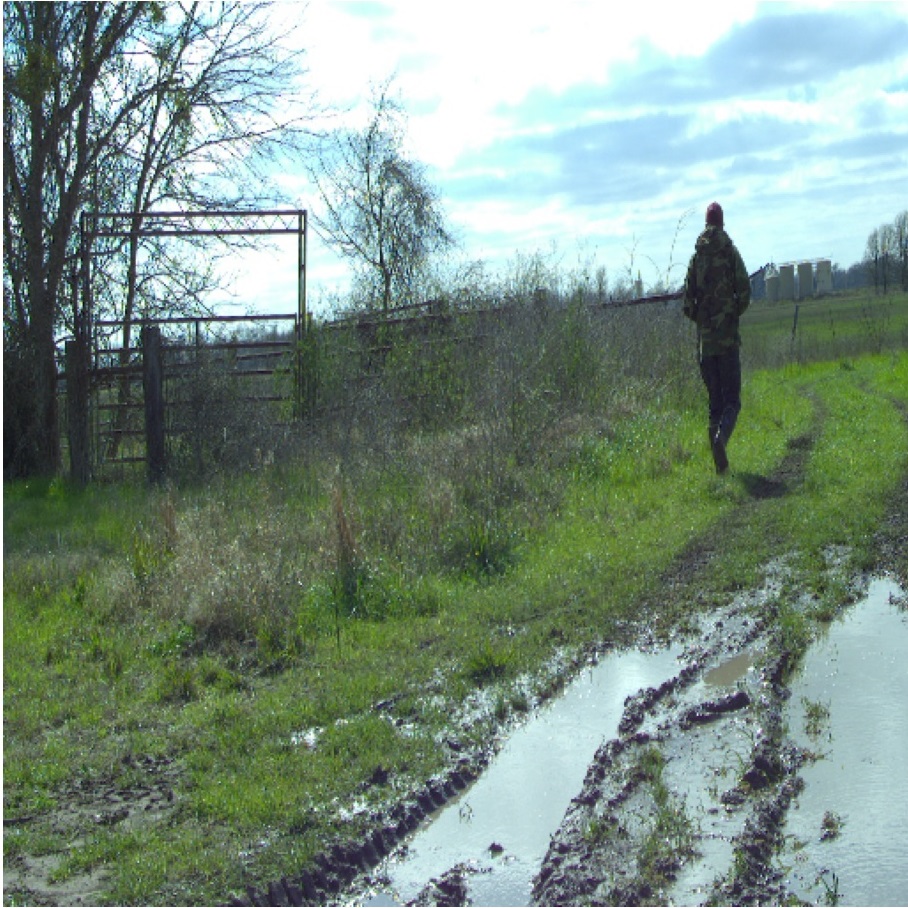} &
        \includegraphics[width=0.175\textwidth]{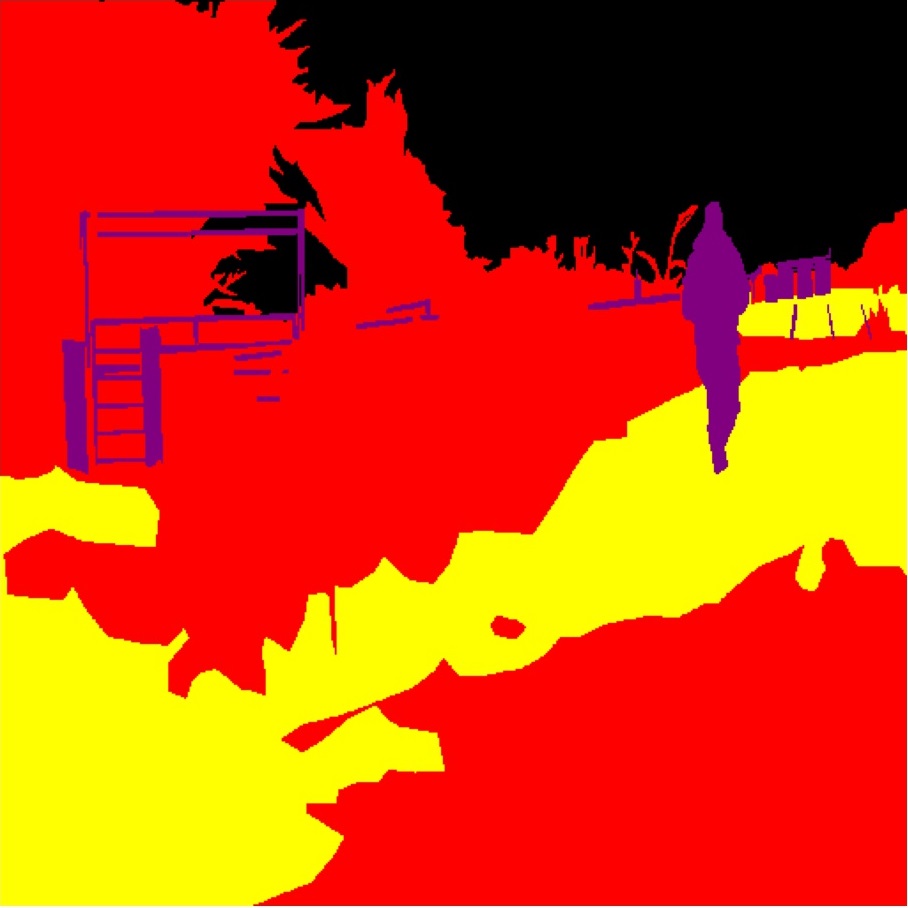} &
        \includegraphics[width=0.175\textwidth]{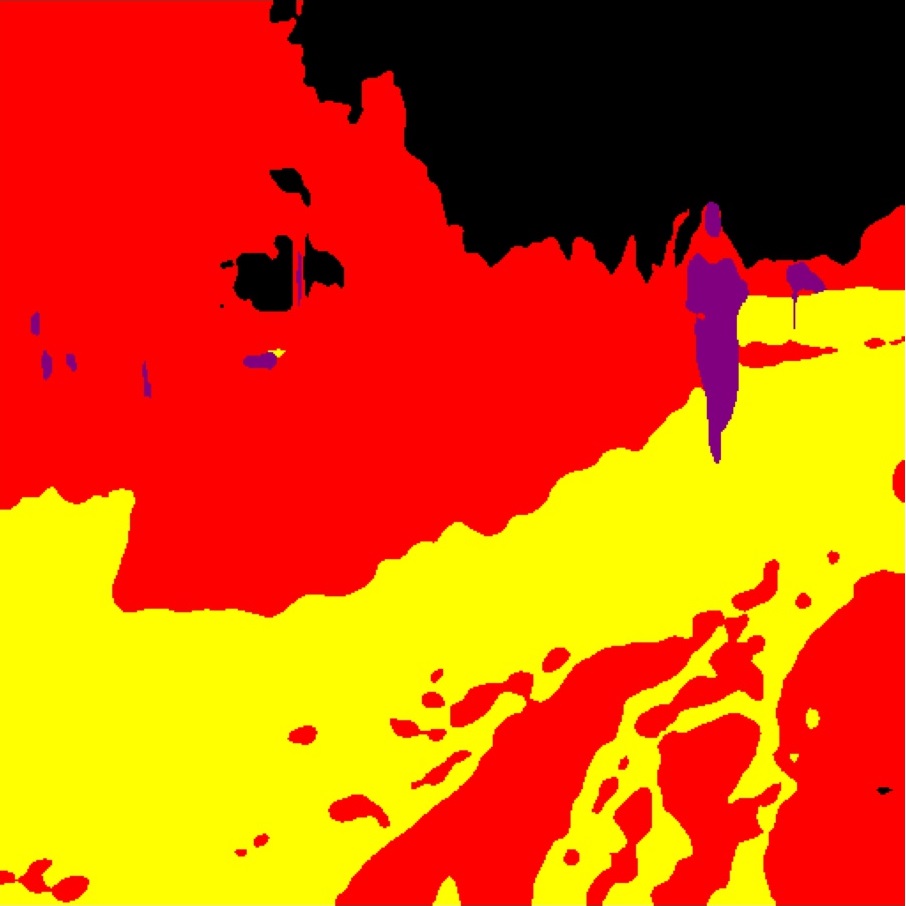} &
        \includegraphics[width=0.175\textwidth]{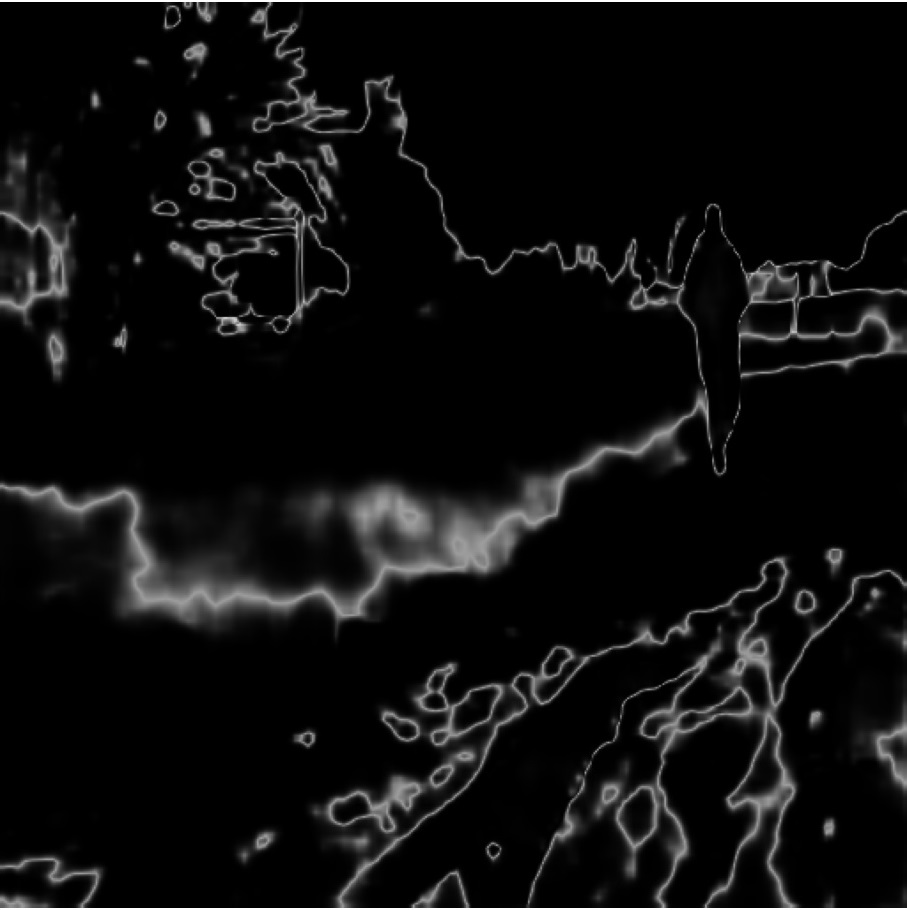} &
        \includegraphics[width=0.175\textwidth]{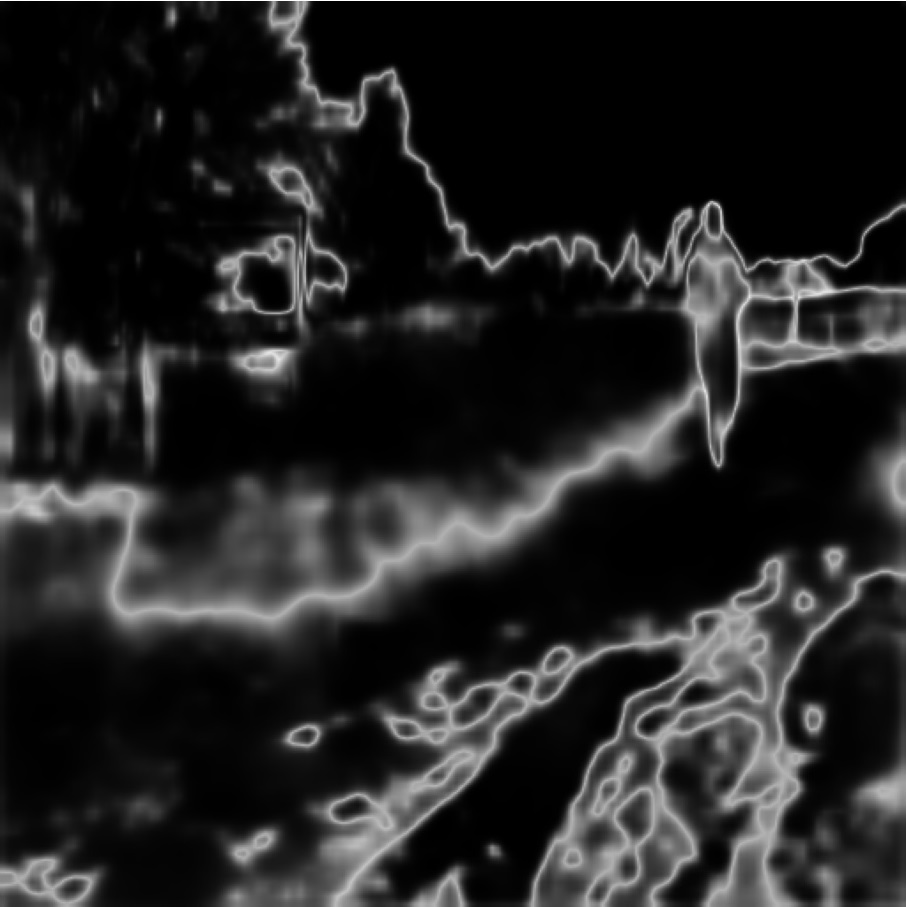} \\

    \end{tabular}
    \caption{In-domain cases from the Rellis3D dataset. The segmentation predictions are often correct (third column) and larger uncertainty values are assigned to unclear boundaries between segments for the standard method (last column) than our method (fourth column).}
     \label{fig:in_domain}
\end{figure*}

In Figure \ref{fig:in_domain} the segmentation results on test images from the Rellis3D dataset are shown. These cases are in-domain. When comparing the results to the ground-truth labels, it can be seen they are mostly correct. The uncertainty estimation for our method and the standard method are also shown. It can be seen that larger uncertainty values are assigned to unclear boundaries between segments for the standard method (last than for our method.

\begin{figure*}
\centering
\begin{tabular}{c c c}
    & uncertainty & uncertainty \\
    test image & ours & standard \\
    \includegraphics[width=0.175\textwidth]{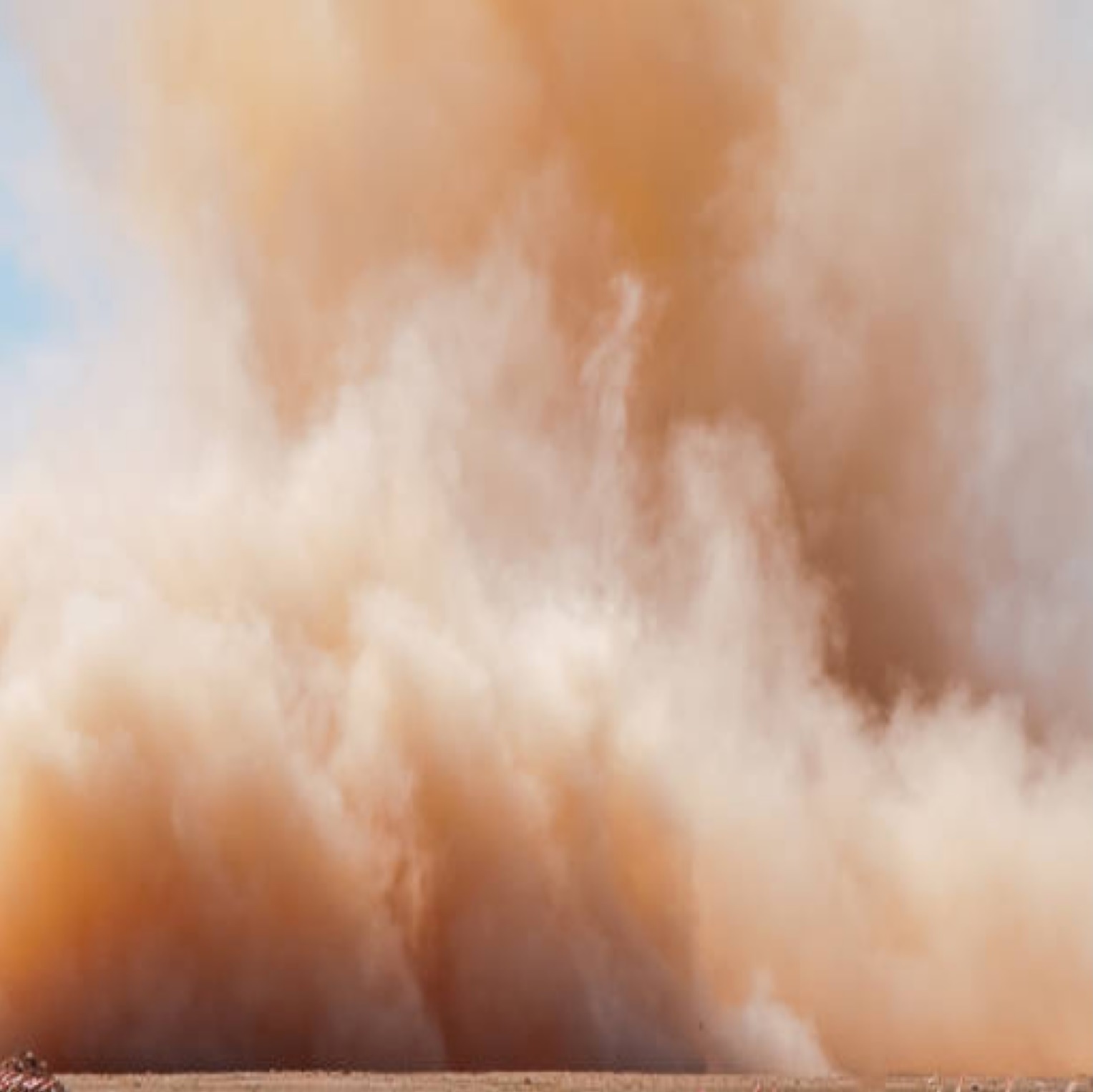}
    &
    \includegraphics[width=0.175\textwidth]{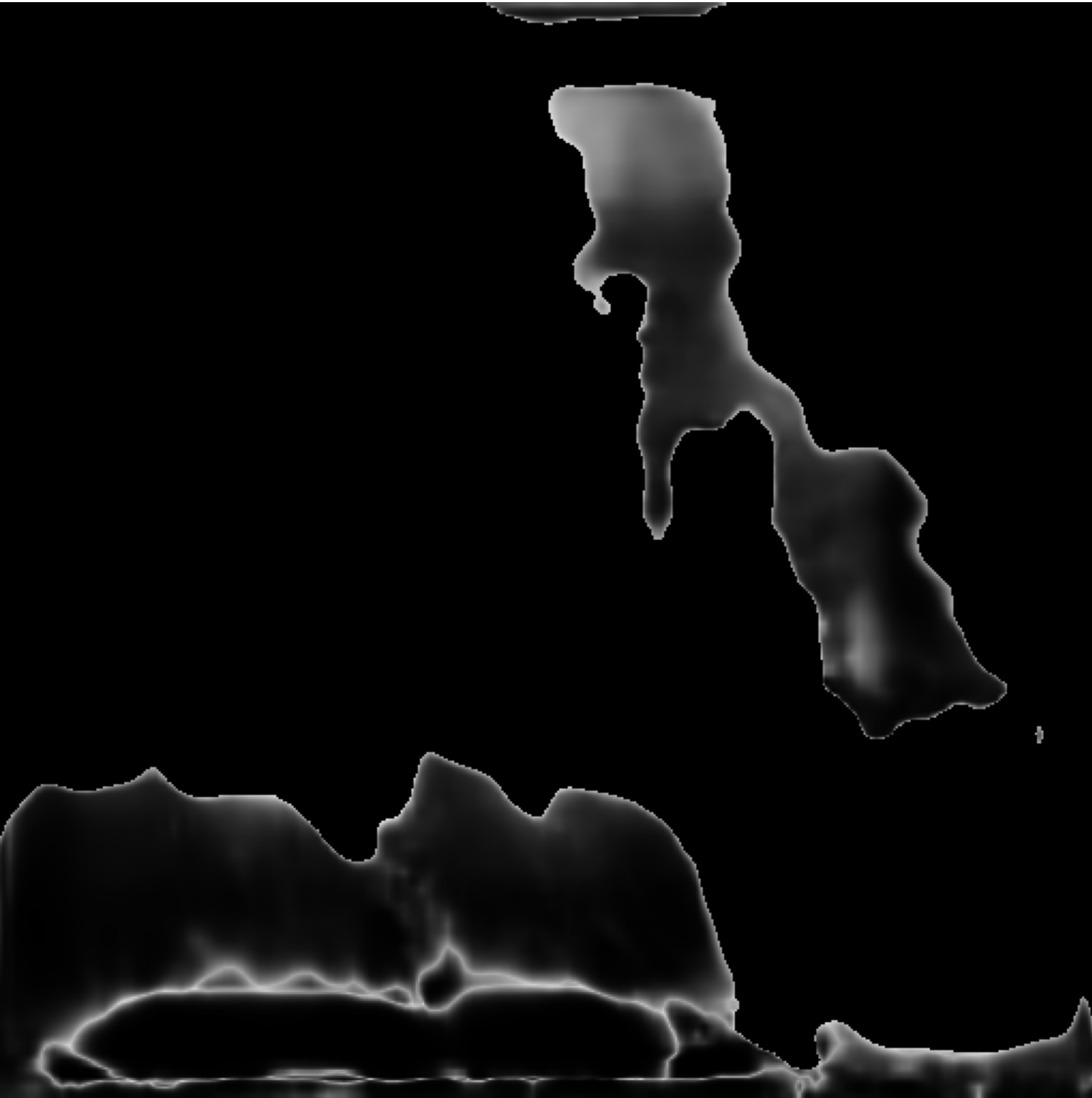}
    &
        \includegraphics[width=0.175\textwidth]{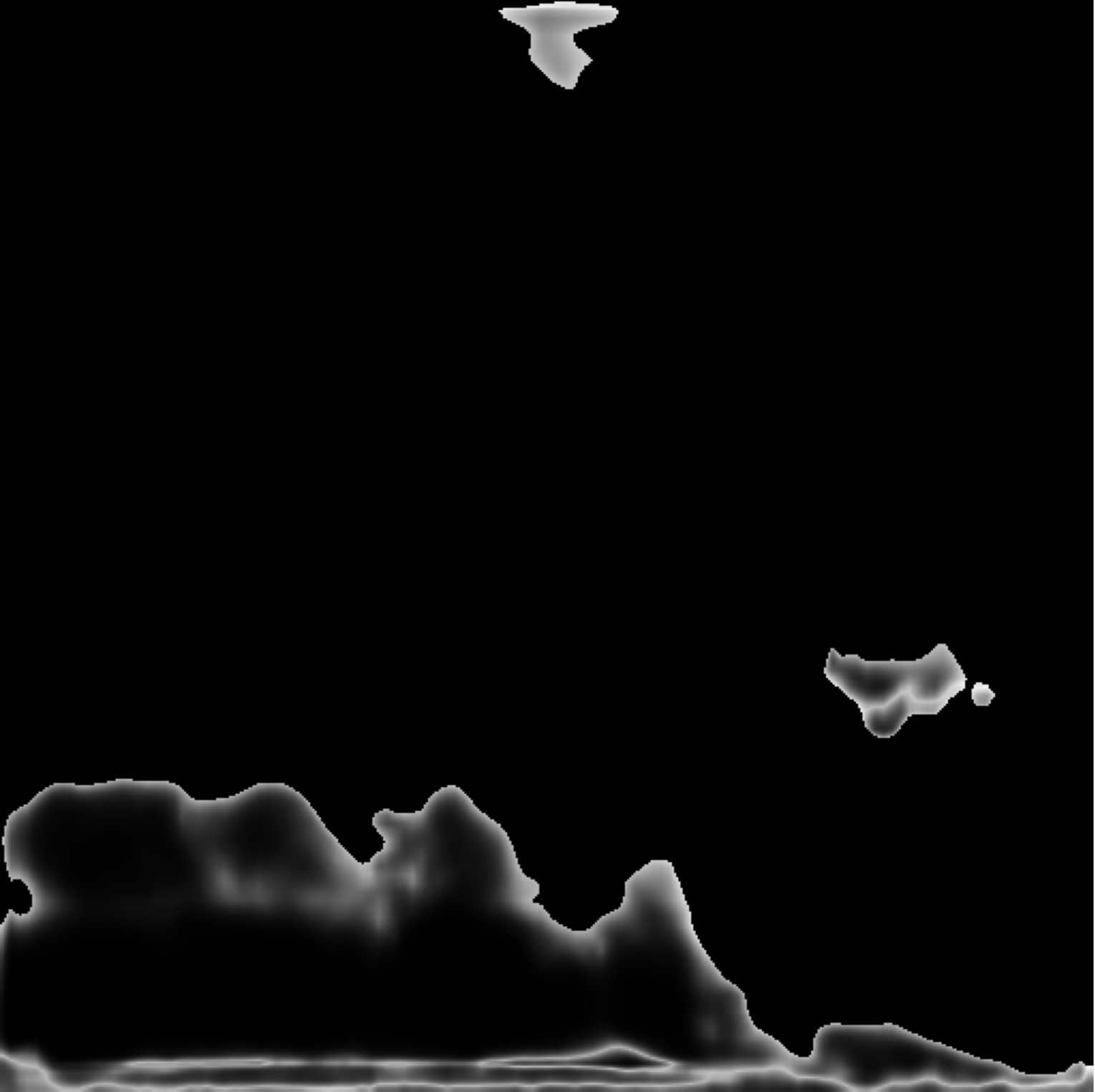}
    \\
   \multicolumn{3}{c}{Fog}\\
   \includegraphics[width=0.175\textwidth]{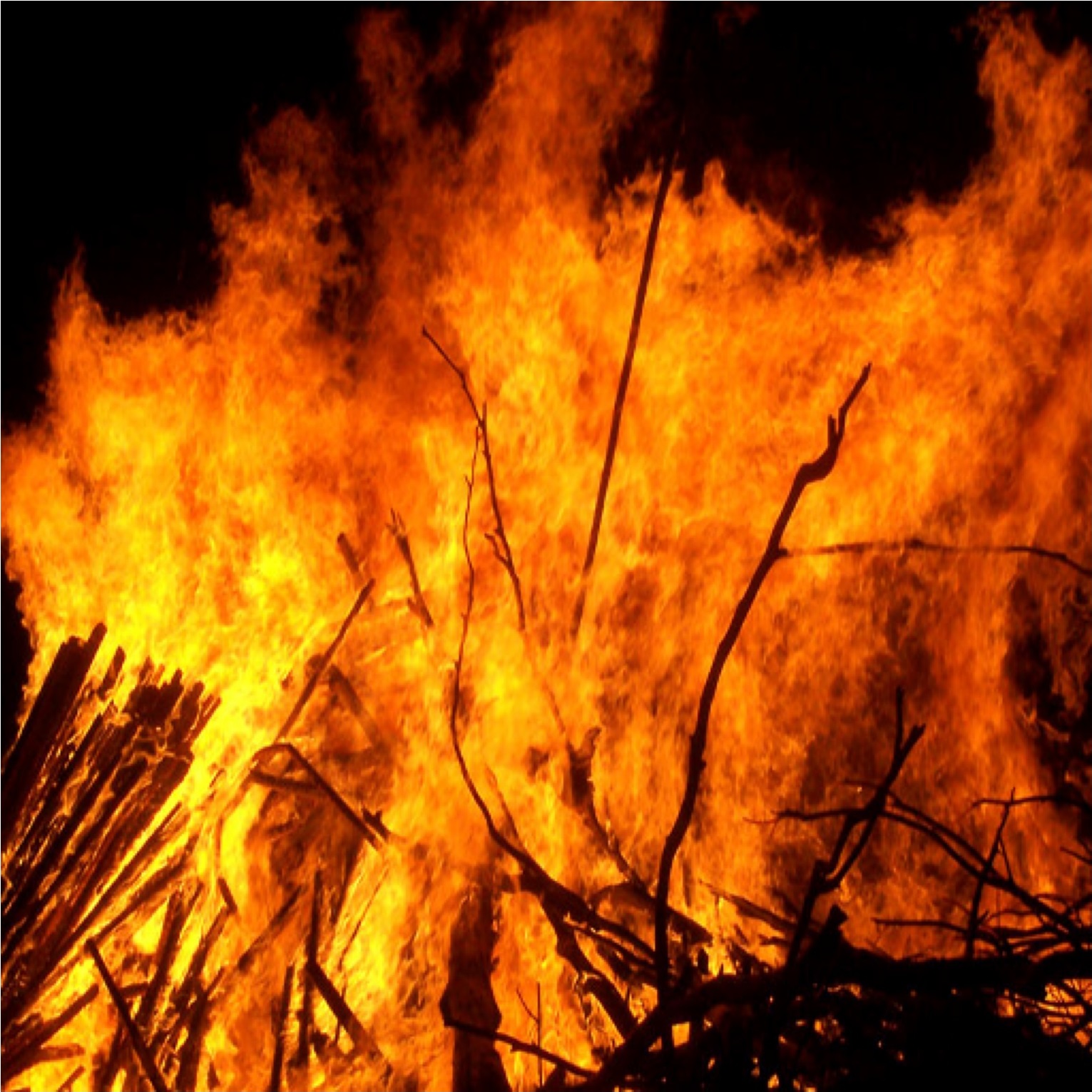}
 &
        \includegraphics[width=0.175\textwidth]{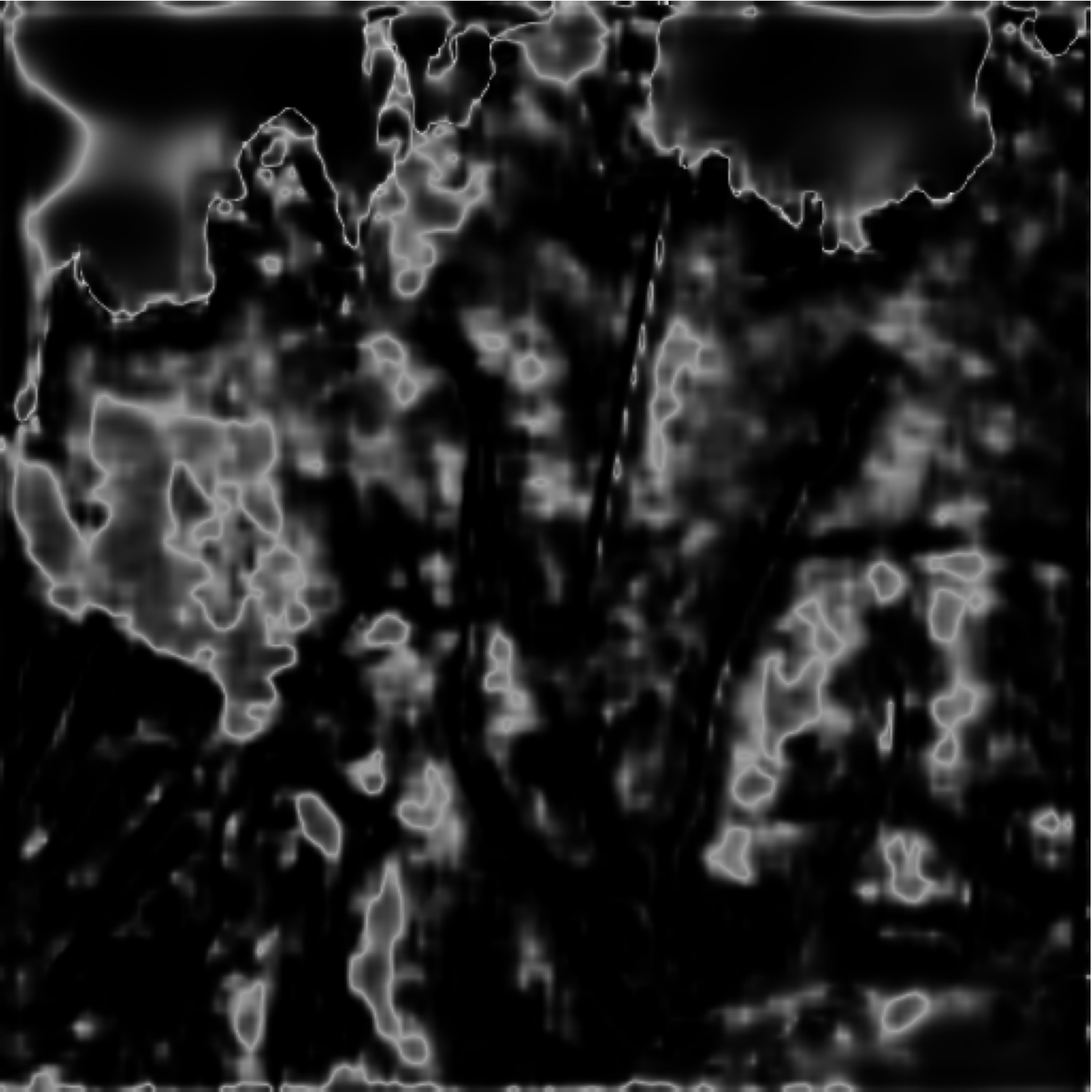}
    &
        \includegraphics[width=0.175\textwidth]{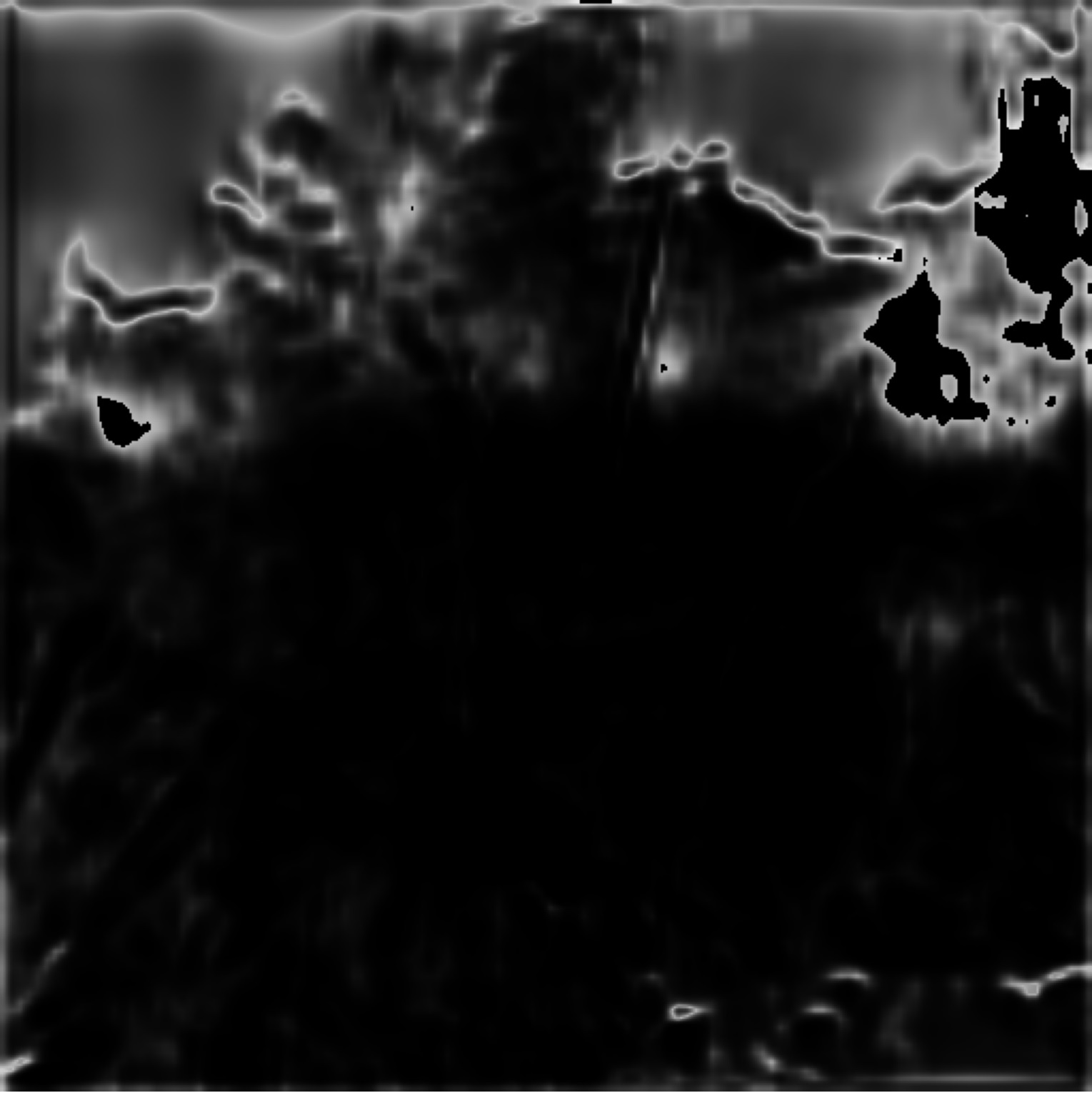}
    \\
     \multicolumn{3}{c}{Fire}\\

    \includegraphics[width=0.175\textwidth]{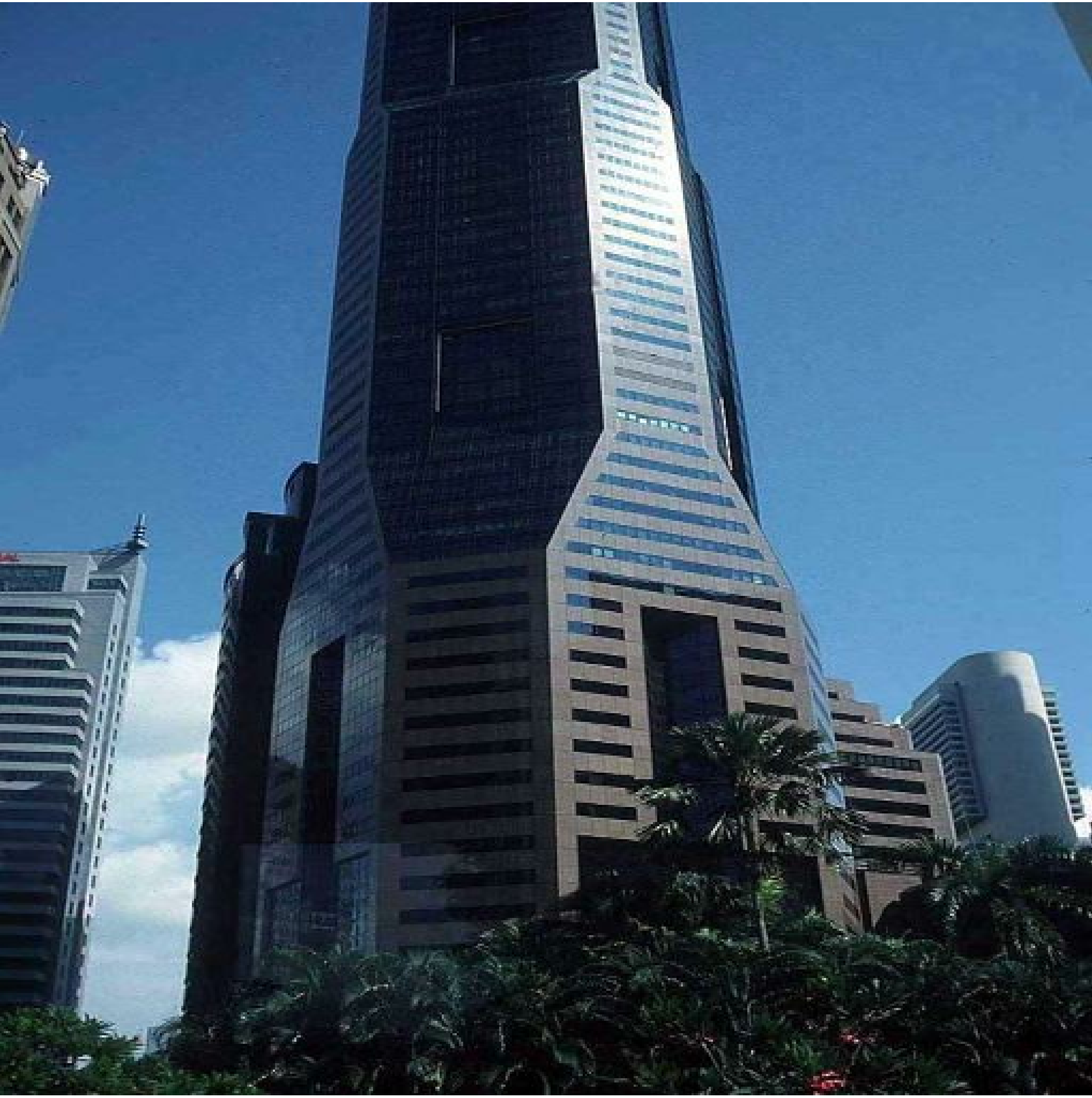}
    &
    \includegraphics[width=0.175\textwidth]{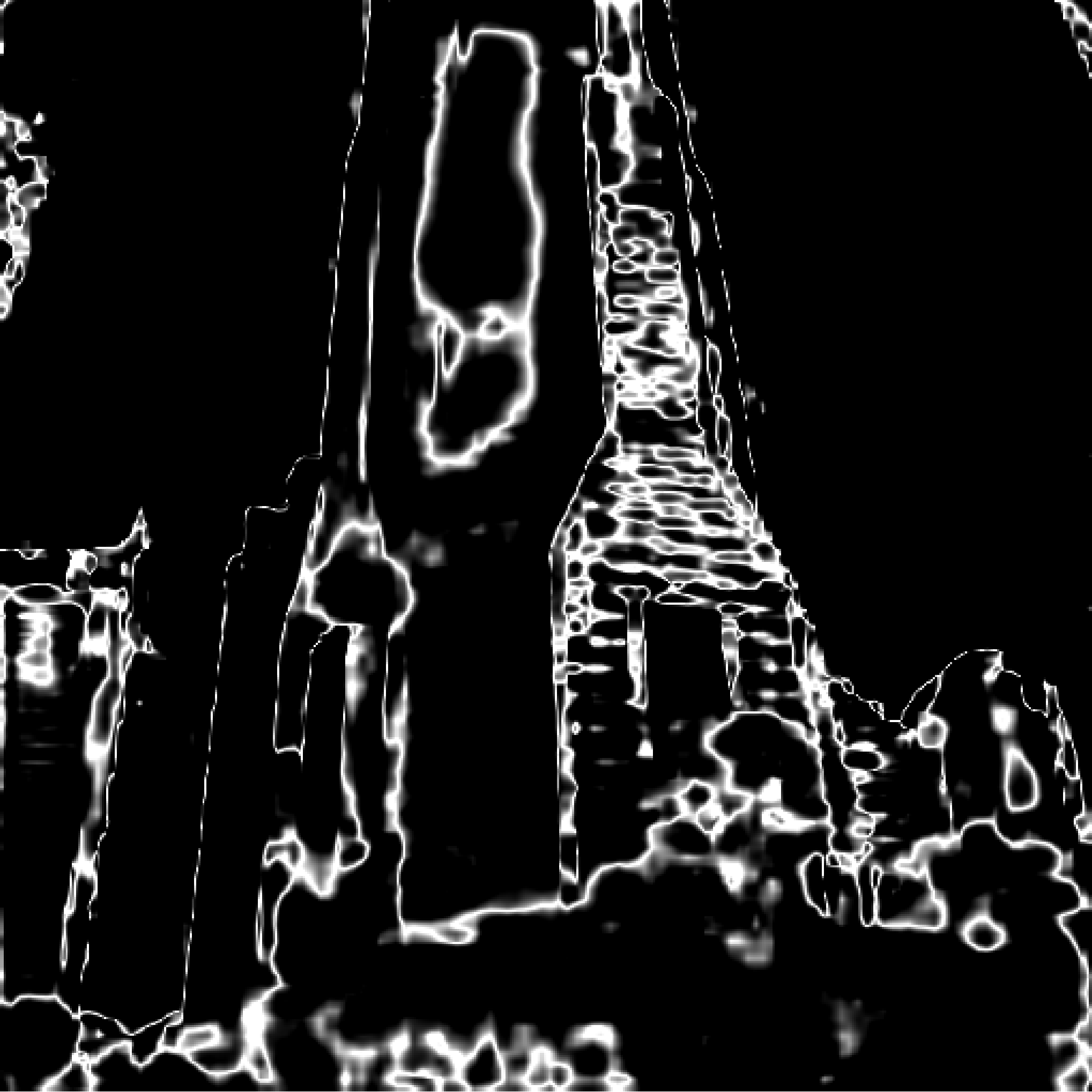}
    &
        \includegraphics[width=0.175\textwidth]{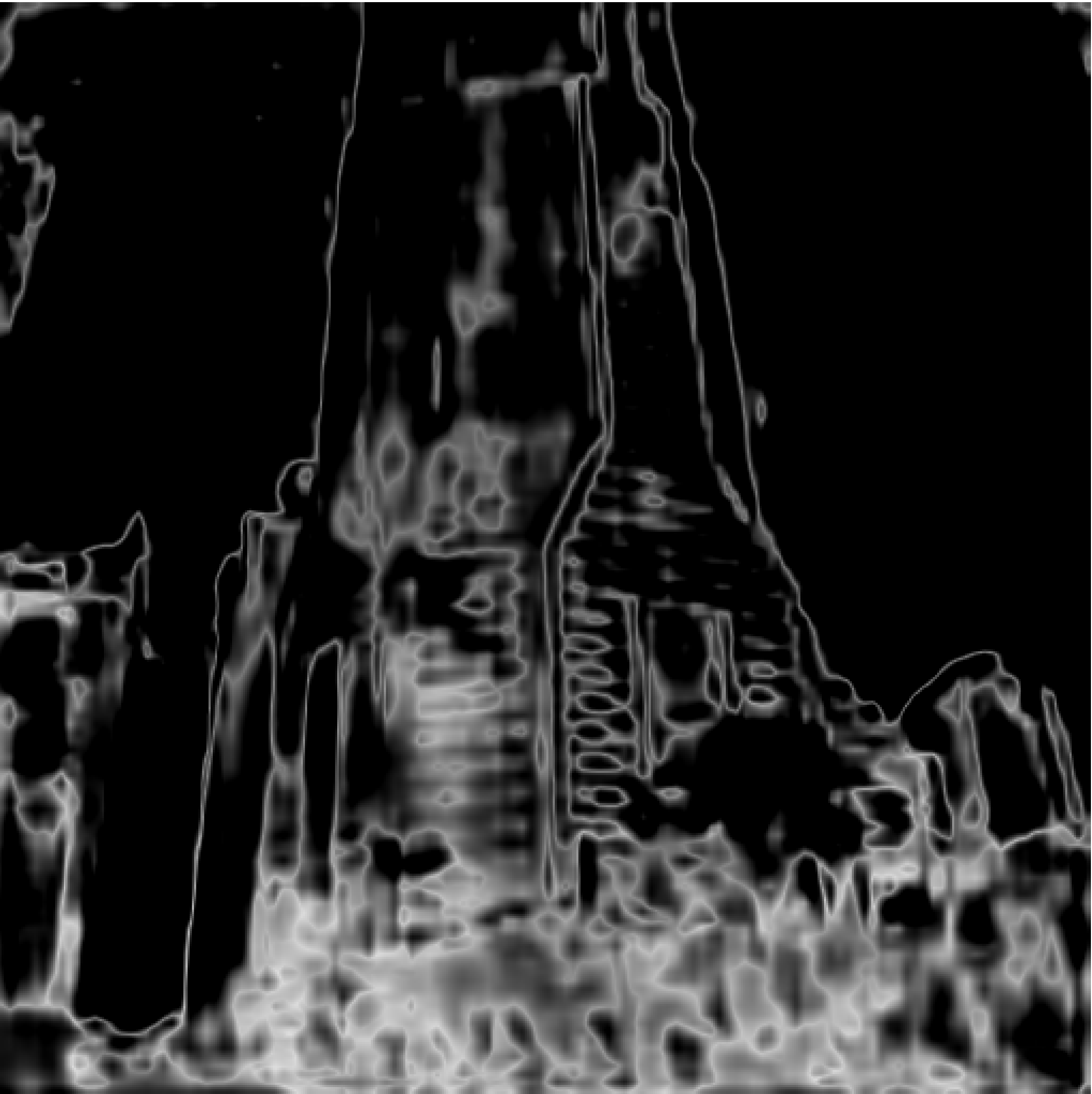}
    \\
   \multicolumn{3}{c}{Building}\\
   \includegraphics[width=0.175\textwidth]{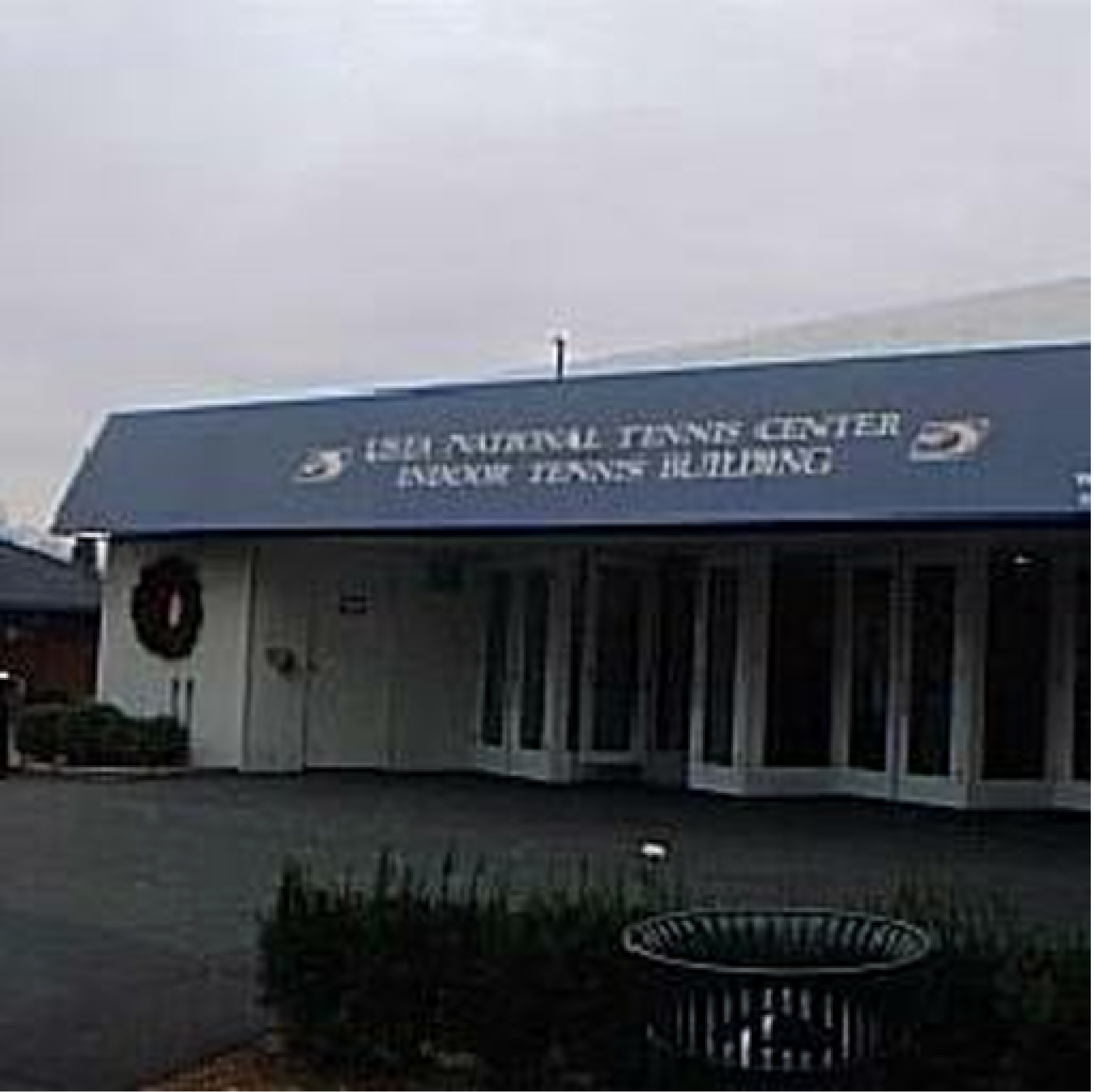}
 &
        \includegraphics[width=0.175\textwidth]{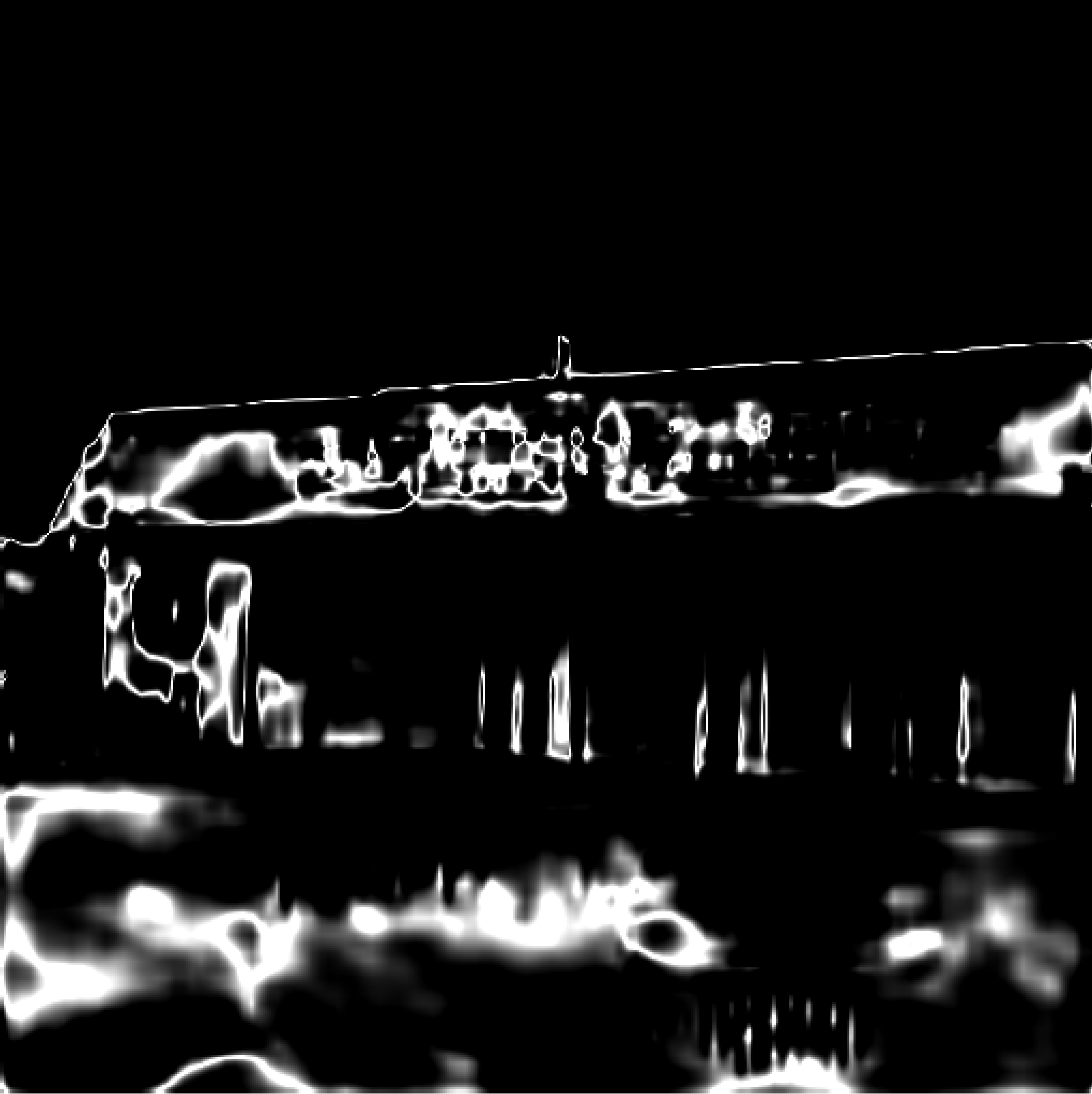}
    &
        \includegraphics[width=0.175\textwidth]{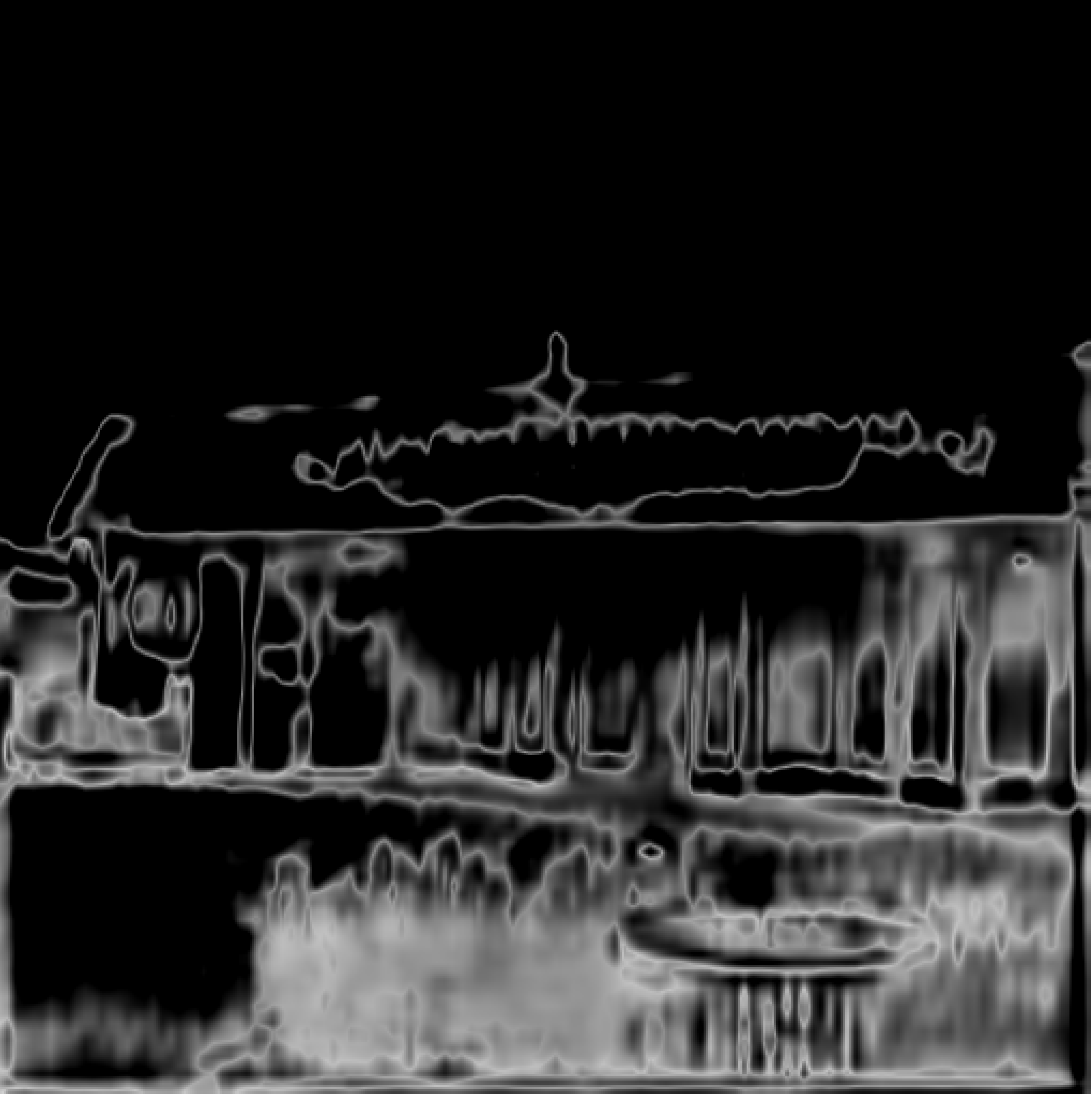}
    \\
   \multicolumn{3}{c}{House}\\
    \includegraphics[width=0.175\textwidth]{figures/OOD_tree_in.png}
 &
        \includegraphics[width=0.175\textwidth]{figures/OOD_tree_ours.png}
    &
        \includegraphics[width=0.175\textwidth]{figures/OOD_tree_standard.png}
    \\
   \multicolumn{3}{c}{Countryside}\\
     \includegraphics[width=0.175\textwidth]{figures/OOD_pigs_in.png}
 &
        \includegraphics[width=0.175\textwidth]{figures/OOD_pigs_ours.png}
    &
        \includegraphics[width=0.175\textwidth]{figures/OOD_pigs_standard.png}
    \\
   \multicolumn{3}{c}{Farm}\\
   \end{tabular}
    \caption{Visualisation of the uncertainty for images with out-of-domain regions: two images from our dataset (fog and fire) and four images from the SceneParse150 \cite{sceneparse150} dataset (building, house and countryside).}
     \label{fig:out_domain}
\end{figure*}

In Figure \ref{fig:out_domain} The uncertainty estimation for our method and the standard method are shown for images with out-of-domain regions: for two images from our own dataset, and four images from the SceneParse150 \cite{sceneparse150} dataset.The uncertainty estimates with our method are higher than those obtained with the standard method.

\begin{figure*}[!t]
    \centering
     \begin{subfigure}[b]{0.45\textwidth}
         \centering
         \includegraphics[width=\textwidth]{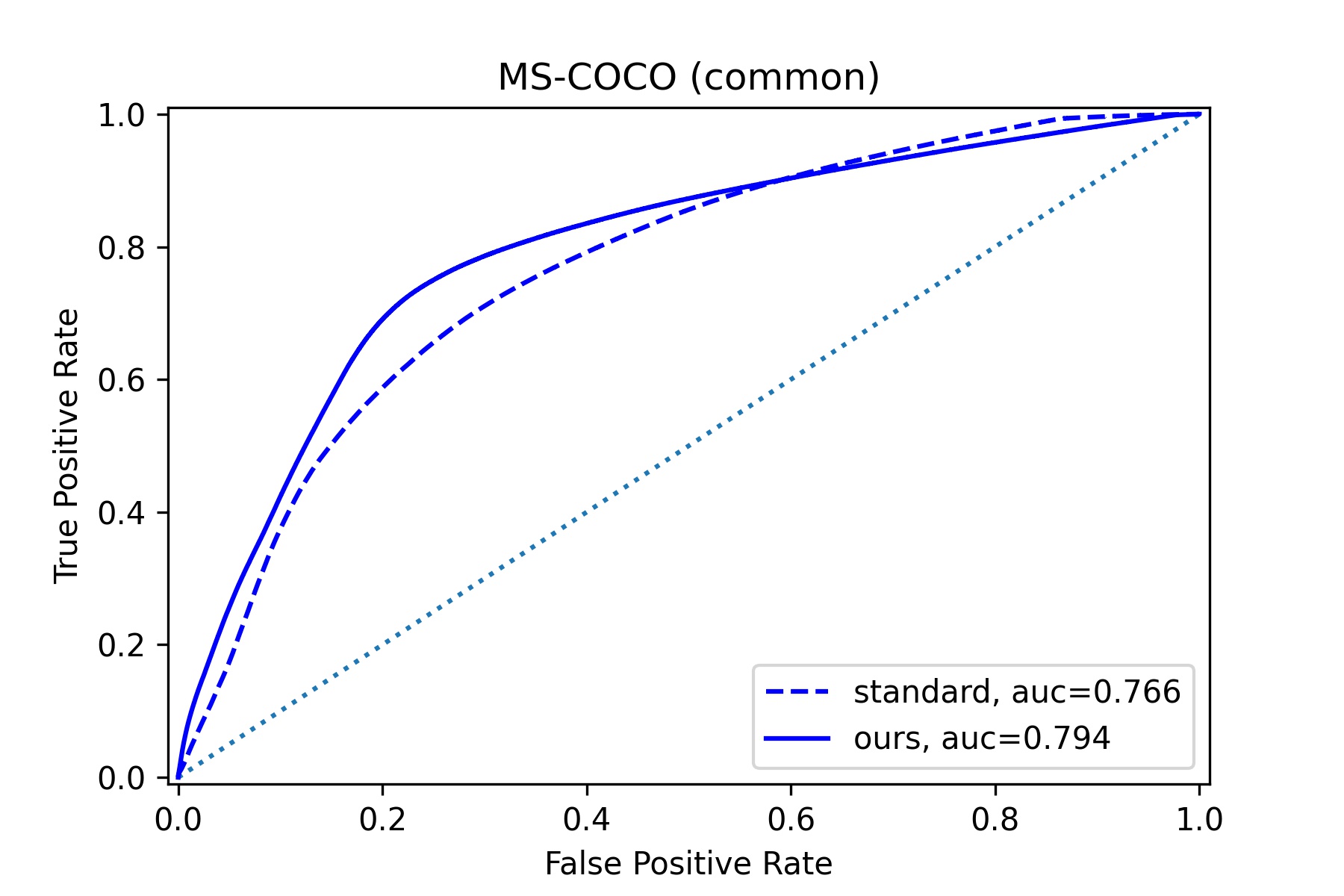}
         \caption{MS-COCO: common objects}
         \label{fig:roc_MS_COCO}
     \end{subfigure}
     \begin{subfigure}[b]{0.45\textwidth}
         \centering
         \includegraphics[width=\textwidth]{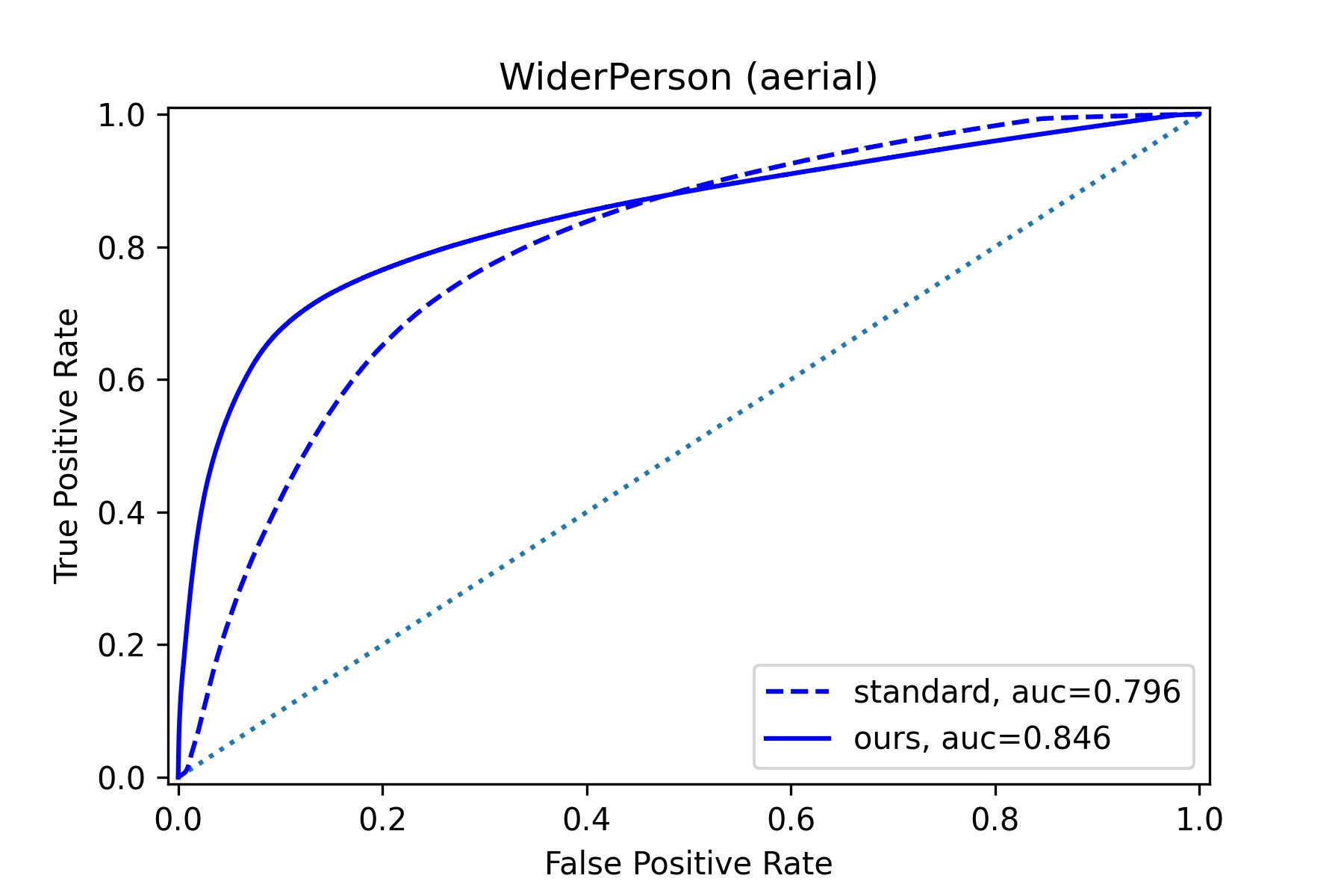}
         \caption{WiderPerson: aerial}
         \label{fig:roc_widerperson}
     \end{subfigure}
     \\
    \centering
      \begin{subfigure}[b]{0.45\textwidth}
         \centering
         \includegraphics[width=\textwidth]{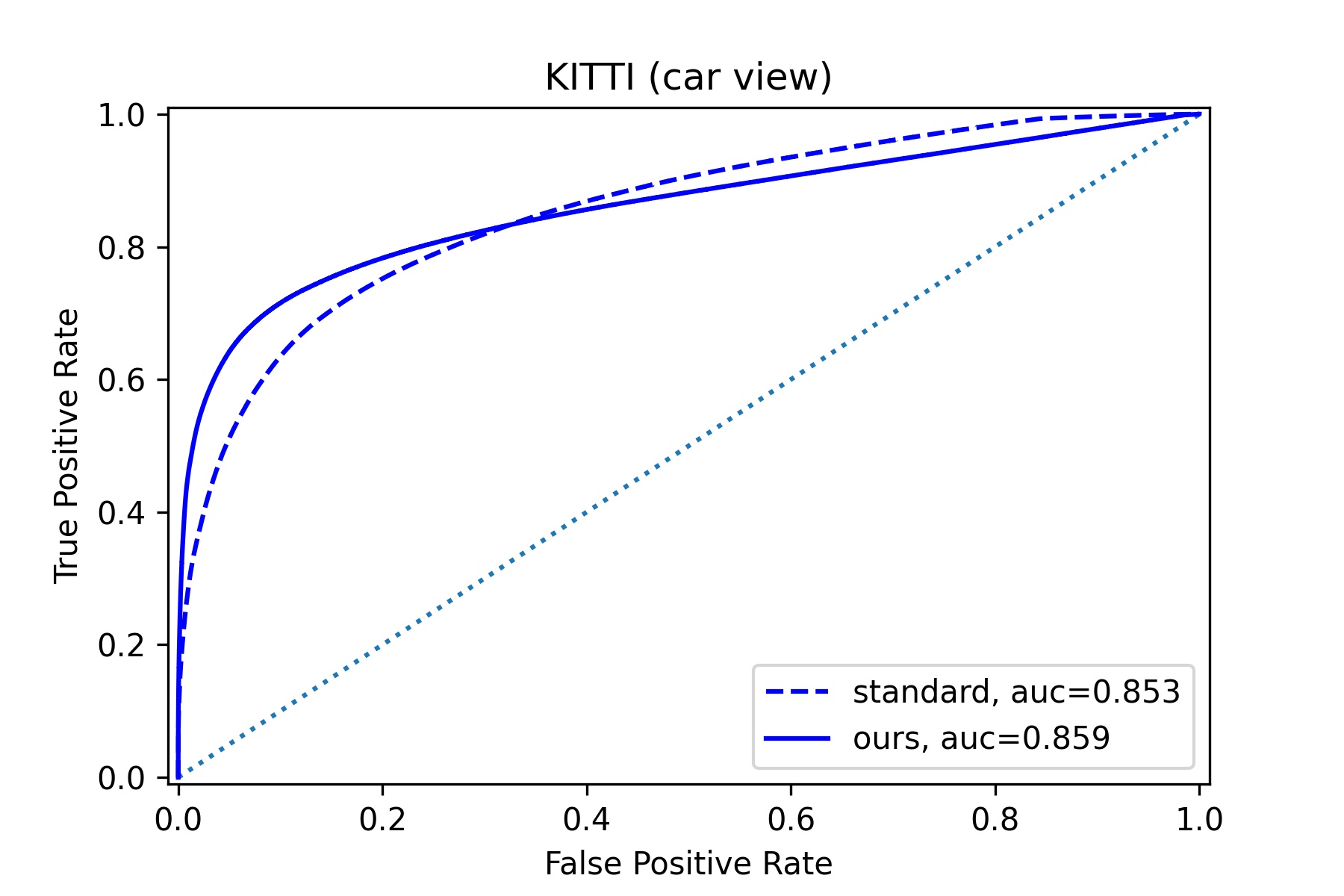}
         \caption{KITTI: self-driving cars}
         \label{fig:KITTI_self_driving_cars}
     \end{subfigure}
       \begin{subfigure}[b]{0.45\textwidth}
         \centering
         \includegraphics[width=\textwidth]{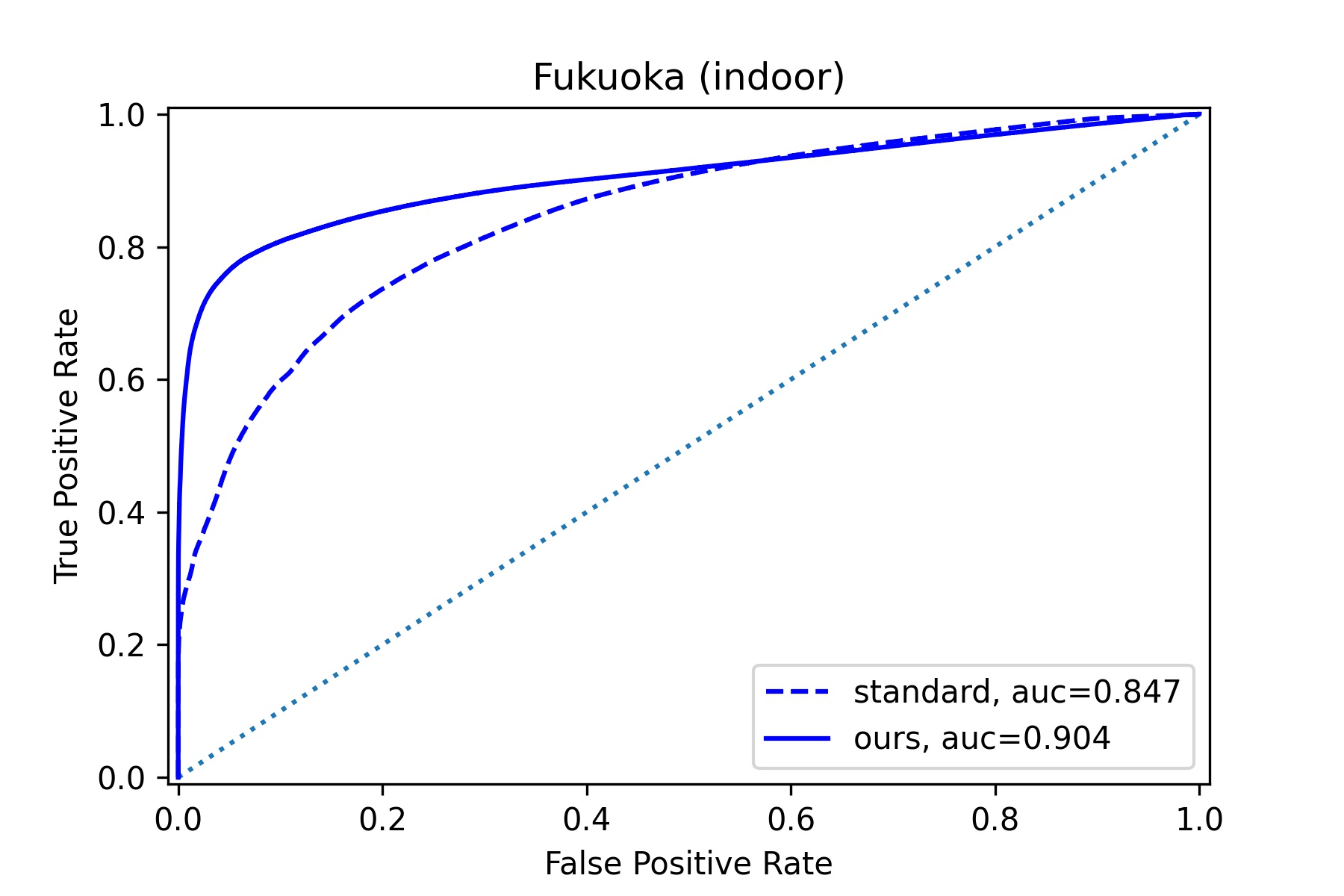}
         \caption{Fukuoka: indoor robot}
         \label{fig:roc_fog_fukuoka_indoor}
     \end{subfigure}
     \\
     \centering
      \begin{subfigure}[b]{0.45\textwidth}
         \centering
         \includegraphics[width=\textwidth]{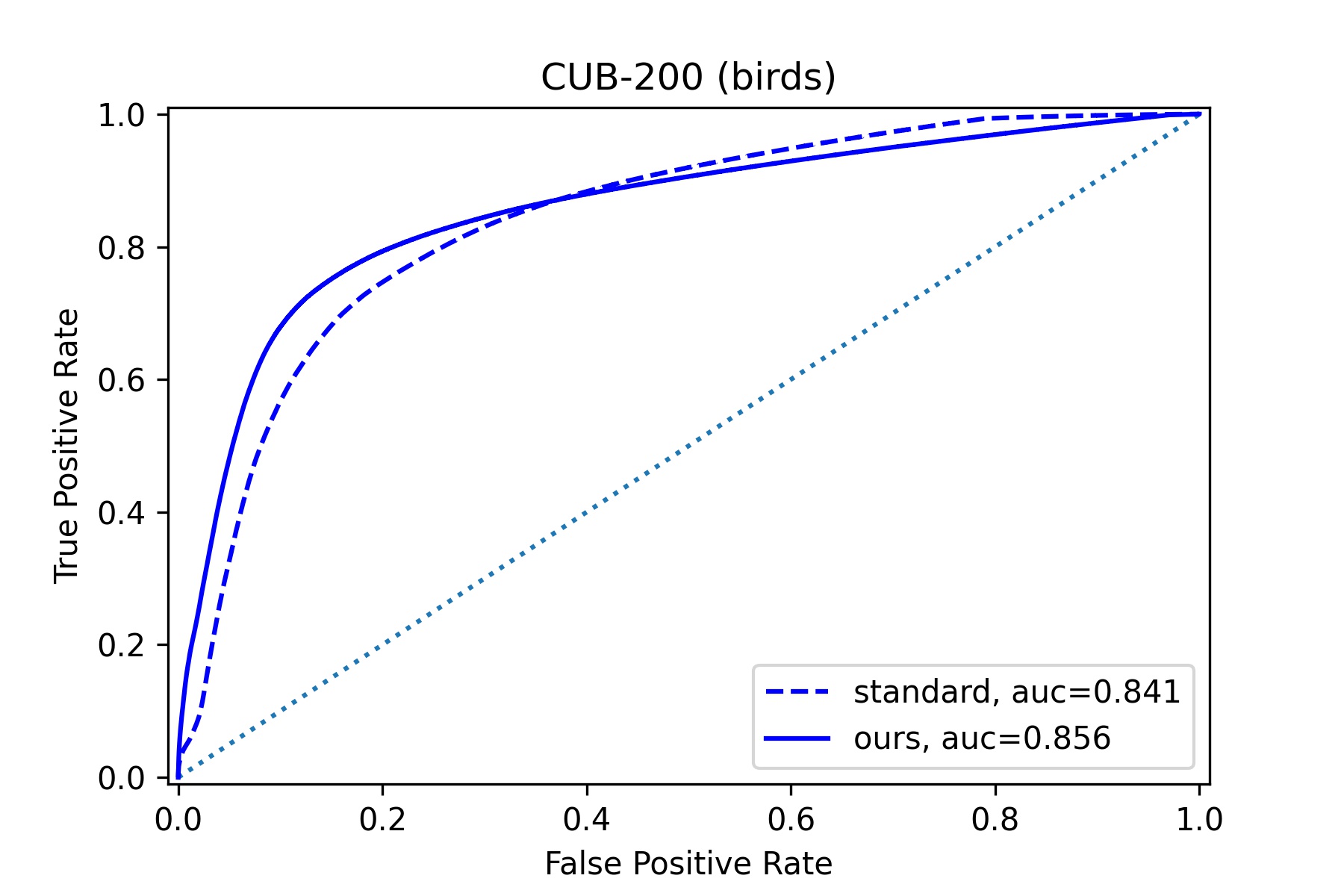}
         \caption{CUB-200: birds}
         \label{fig:roc_CUB_200_birds}
     \end{subfigure}
     \begin{subfigure}[b]{0.45\textwidth}
         \centering
         \includegraphics[width=\textwidth]{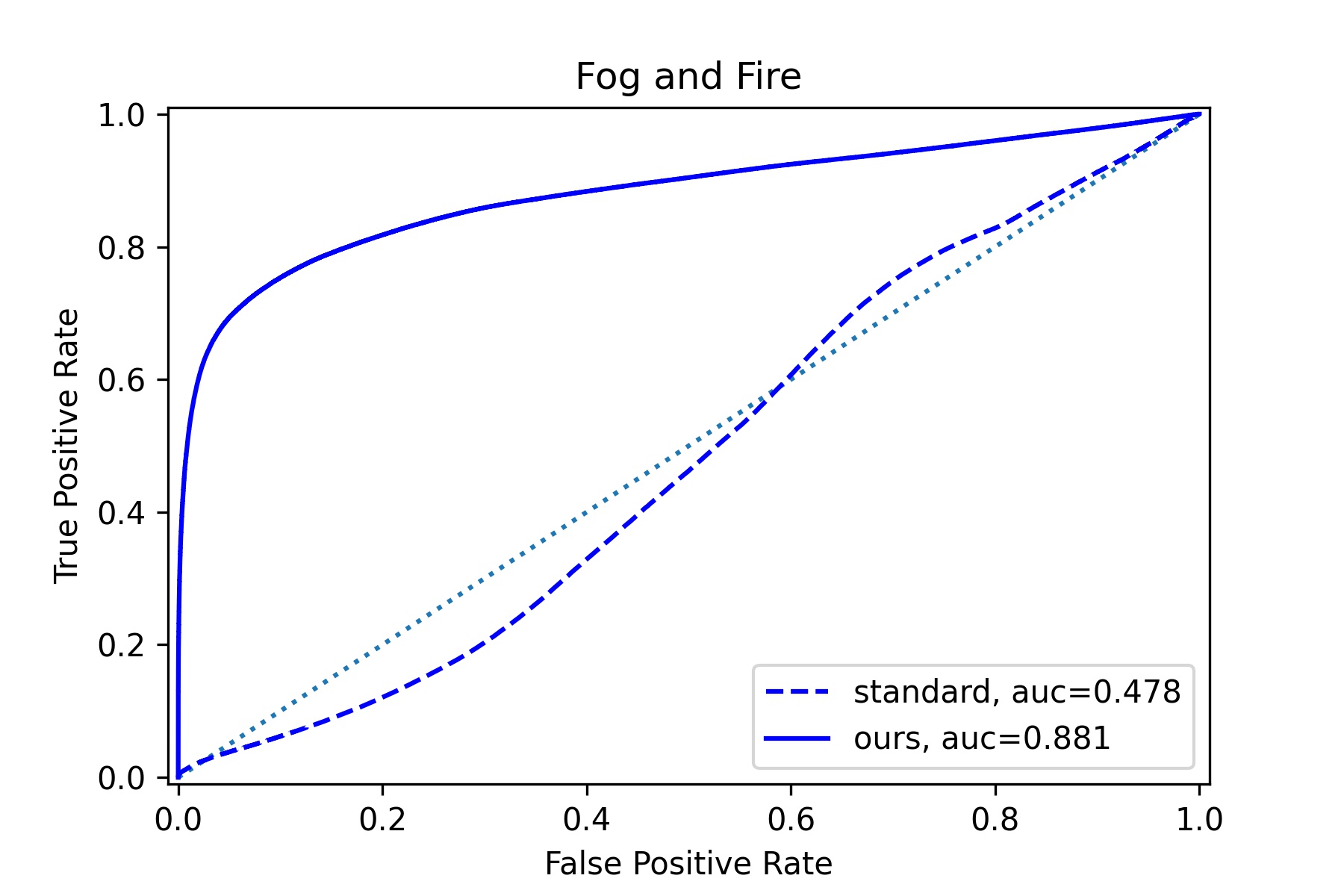}
         \caption{Fog and fire: anomalies}
         \label{fig:roc_fog_fire}
     \end{subfigure}
    \caption{Out-of-distribution performance of our uncertainty estimation on various datasets. The AUC values are also reported in Table~\ref{table:uncertainty_results}}
     \label{fig:roc}
\end{figure*}

In Figure~\ref{fig:roc} the ROC curves for six datasets are shown. These graphs show that our method shows a better uncertainty estimation performance than standard method.

\subsection{Segmentation results and accuracy}

To evaluate the segmentation results of our proposed method, we compare the results with two state-of-the-art methods: DeepLabV3+ as our method is added on top of this method and GA-Nav \cite{ganav}.
We report the Intersection over Union (IoU) for each class $i$ and the mean IoU (mIoU).

Table \ref{table:segmentation_results} shows that our method performs on par with GA-Nav-r8. Surprisingly, our model performs better than the reported DeepLabV3+ in \cite{ganav}, which has the same architecture and setup as model, but was trained with other settings. We hypothesize that some of our design choices are better suited for Rellis3D. One of the differences with \cite{ganav} is that they train with a smaller batch size (2 instead of 4). Their image size is 375 $\times$ 600 pixels, which corresponds less with the aspect ratio of the images in the dataset (1600 $\times$ 1920 pixels) than our image size (512 $\times$ 512 pixels). Possibly the biggest advantage is that we use more augmentations. Where \cite{ganav} uses only horizontal flip and random crop, we add scale, rotate and color jitter. Especially the scale changes in the dataset are sometimes large, which we address by the scale augmentation. Some of the images are recorded under a slight tilt, because the robot might be on a slope. This is addressed with rotation augmentations during training. Examples of the segmentation images are shown in the Appendix in Figure~\ref{fig:in_domain}.

\begin{table}[!t]
\centering
\caption{Segmentation results for the Rellis3D dataset~\cite{rellis} in mIoU for all classes for the DeepLabV3+ \cite{ganav,deeplabv3plus} model, the GA-Nav-r8 \cite{ganav} model and our approach. }
\label{table:segmentation_results}
\begin{tabular}{l cccccc c}
\toprule
    Model & \rotatebox[origin=c]{90}{smooth} & \rotatebox[origin=c]{90}{rough} & \rotatebox[origin=c]{90}{bumpy} & \rotatebox[origin=c]{90}{forbidden} & \rotatebox[origin=c]{90}{obstacle} & \rotatebox[origin=c]{90}{background} & \rotatebox[origin=c]{90}{mean} \\
    \midrule
    DeepLabV3+ & 65.8  & 79.8 & 19.7 & 47.5    &  64.9    & 95.9    &   62.3 \\
    GA-Nav-r8 & 78.5 &  88.3 & 37.3 & 72.3    &  74.8  &   96.1   &    74.4  \\
    \midrule
    \rowcolor{Gray}
    Our method & 84.7 &  86.8 & 59.9 & 80.2  &    43.5   &  97.0   &    \textbf{75.3} \\
    \bottomrule
\end{tabular}
\vspace{-4mm}
\end{table}

\end{document}